\documentclass{article}
\PassOptionsToPackage{numbers,square,sort&compress}{natbib}

\usepackage[preprint]{neurips_2026}

\usepackage[utf8]{inputenc} 
\usepackage[T1]{fontenc}    
\usepackage{hyperref}       
\usepackage{url}            
\usepackage{booktabs}       
\usepackage{amsfonts}       
\usepackage{nicefrac}       
\usepackage{microtype}      
\usepackage{xcolor}         
\usepackage{graphicx}
\usepackage{subcaption}
\usepackage{enumitem}
\usepackage{amsmath,amssymb,amsfonts}
\usepackage{booktabs}
\usepackage{array}
\usepackage{tabularx}
\usepackage{graphicx}
\usepackage{caption}
\usepackage{subcaption}
\usepackage{xcolor}
\usepackage{microtype}
\usepackage{enumitem}
\usepackage{float}
\usepackage{hyperref}
\usepackage[capitalize,nameinlink]{cleveref}
\usepackage[normalem]{ulem}
\usepackage{placeins}

\crefname{appendix}{Appendix}{Appendices}
\Crefname{appendix}{Appendix}{Appendices}

\graphicspath{{papers/iclr2027-causal-statehood/}{./}}

\definecolor{deepteal}{RGB}{20,92,84}
\definecolor{controlgray}{RGB}{78,82,88}
\definecolor{alertcoral}{RGB}{201,78,61}
\definecolor{softblue}{RGB}{67,126,171}
\definecolor{lightteal}{RGB}{223,239,236}
\definecolor{lightcoral}{RGB}{249,225,220}
\definecolor{lightamber}{RGB}{250,222,140}
\definecolor{addedmaterialred}{RGB}{166,45,45}
\newcommand{\hidphase}[1]{\colorbox{lightteal}{\strut #1}}
\newcommand{\editsite}[1]{\colorbox{lightamber}{\strut #1}}
\newcommand{\CUS}{\mbox{CUS}}
\newcommand{\msp}{m\cdot\tilde{s}}
\newcommand{\ms}{m\cdot s}
\newcommand{\hnext}{\hat{h}_{t+1}}
\newif\ifshowaddedmaterial
\showaddedmaterialtrue

\newcommand{\addedtext}[1]{%
  \ifshowaddedmaterial\textcolor{addedmaterialred}{#1}\else#1\fi
}
\allowdisplaybreaks

\hypersetup{colorlinks=true,linkcolor=softblue,urlcolor=softblue,citecolor=softblue}
\crefname{section}{Section}{Sections}
\Crefname{section}{Section}{Sections}
\crefname{appendix}{Appendix}{Appendices}
\Crefname{appendix}{Appendix}{Appendices}
\title{When Does Activation Steering Change What a Model Computes From?}

\author{%
Benjamin Shih \quad John Winnicki \quad Eric Darve
\\
Institute for Computational and Mathematical Engineering
\\
Stanford University
}

\begin{document}
\maketitle

\begin{abstract}
Activation steering can reliably change an agent’s output by modifying its internal activations. 
Yet arriving at the same answer need not involve the same computation: behavioral equivalence does not imply mechanistic equivalence.
We test whether an activation edit changes the state used in subsequent computation or biases that computation toward the desired output.
In a controlled state-tracking task, trace supervision creates an editable register: changing its internal value makes the model apply the next operation to the edited state.
We next ask whether widely used mean activation steering admits a similar interpretation: does steering recreate the internal configuration the model naturally uses when performing a task? 
If so, transplanting that activation should be effective, and steering should remain effective at the scale of the natural source-to-target activation change. 
Neither prediction holds in the two selected model–task settings.
Replacing the activation at one layer produces less than 2\% of the target-answer margin gain from patching through all remaining layers, while natural-scale steering is similarly ineffective. 
The steering vectors have norms 77 and 26 times the median natural change in Qwen and Llama, producing 54\% and 82\% of the reference effect.
Thus, a successful steering intervention need not reproduce the natural target activation at the intervention layer.
More generally, an intervention should be interpreted as changing the computational state only when a later computation uses the edited value according to the semantics of that state.
\end{abstract}

\section{Introduction}

Activation steering changes the behavior of an agent by adding a direction to its internal activations \cite{turner2023actadd,todd2024fv,zou2023repe}. 
The direction is often constructed by averaging activations associated with a task or behavior. 
Successful steering may suggest that the intervention writes a task state the model subsequently uses in computation.
On the other hand, it could also simply be biasing later computation towards the same output.
Distinguishing these explanations requires testing how subsequent computation uses the edited activation.


We fine-tune multiple 7B language models to write the running state of a known transition system. 
At evaluation, we change the internal representation of one written state while leaving the visible scratchpad unchanged. 
Because the update rule is known, a genuine state change has a unique downstream consequence: the model should apply the next operation to the edited state rather than the original one. 
The resulting continuation follows the edited state and changes appropriately when the next operation changes, showing that the model is computing from the edited value rather than simply producing a different answer. 

\begin{figure}[t]
\centering
\begin{subfigure}[t]{0.55\linewidth}
\centering
\includegraphics[width=\linewidth]{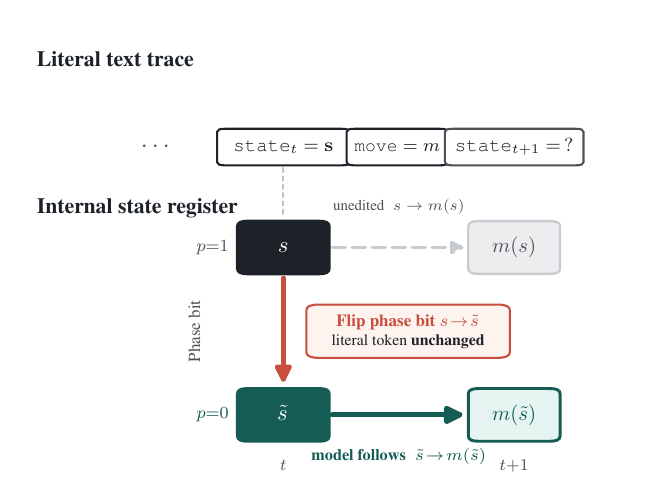}
\caption{The editing mechanism}\label{fig:hero-a}
\end{subfigure}\hfill
\begin{subfigure}[t]{0.43\linewidth}
\centering
\includegraphics[width=\linewidth]{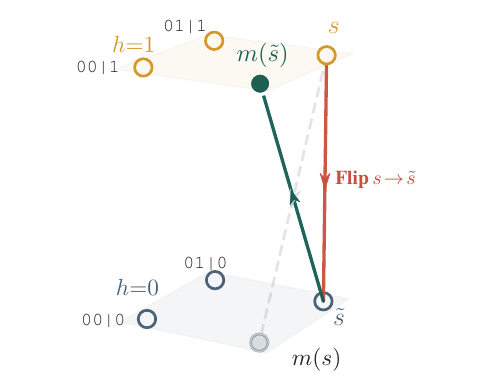}
\caption{A sample divergence from the edit}\label{fig:hero-b}
\end{subfigure}
\caption{\textbf{Editing the register redirects the computation.} The visible trace token reports
state (s), but a low-rank internal edit changes its hidden state to \(\tilde{s}\). Under the same
upcoming move \(m\), the model computes \(m(\tilde{s})\) rather than \(m(s)\). The outcome therefore
depends on both the edited state and the operation applied after the edit.}
\label{fig:hero}
\end{figure}

However, this result does not imply that every behavior-changing edit works this way.
We next test the \emph{correspondence hypothesis (CH)} that Function Vectors \cite{todd2024fv}, a widely used form of mean activation steering, recreate the activation the model naturally produces at the intervention layer when performing the target task.
If CH holds, the target-run activation should be effective when transplanted into the source run, and the steering vector should remain effective when scaled to the natural activation difference between the runs. 

We find that neither prediction holds in the two tested settings. 
These results separate three properties often conflated in steering work: behavioral efficacy, directional specificity, and state like use. 
A steering vector can exhibit the first two without establishing the third. 

\paragraph{Our contributions are as follows:}
\begin{enumerate} 
\item \textbf{A causal test for computational state.} We distinguish changing an answer, changing later computation, and changing a state variable by asking what future operations the edited activation can support.
\item \textbf{A positive state-editing example.} Counterfactual register editing shows that trace supervision can make an explicit scratchpad token into internal state that the model reads and updates. 
Controls separating state from answer replicate in a second 7B model family.
\item \textbf{A natural-activation test for mean steering.} In Qwen and Llama, a large steering vector changes behavior while the matched target activation at one layer and a natural-sized steering edit do not. 
Steering success therefore does not by itself establish that a natural target state has been transplanted.
\item \textbf{A model-dependent specificity result.} Llama responds selectively to the task's steering direction, while Qwen responds to several structured directions of the same size. 
The specificity of a steering vector must be measured rather than assumed from its construction.
\end{enumerate}
\section{Related work}
\label{sec:related}

Activation steering methods, including Activation Addition, representation engineering, inference-time intervention, and Function Vectors, show that internal directions can reliably control language-model behavior \cite{turner2023actadd,zou2023repe,li2023iti,todd2024fv}. Because behaviorally equivalent interventions need not identify a unique internal representation \cite{venkatesh2026nonidentifiability}, we ask whether a successful steering vector recreates an activation that the model naturally uses. This connects to causal abstraction and interchange interventions \cite{geiger2024das,geiger2021abstractions} and to reusable-interface methods such as Hidden APIs \cite{ma2026hiddenapis}. Our controlled register setting makes the state and transition rule known, allowing each edit to be tested through its operation-specific downstream consequence. A fuller discussion of activation patching, scratchpads, state tracking, world models, and representation geometry appears in \Cref{app:related-work}.

\section{A controlled scratchpad state-tracking task}
\label{sec:task}

The model reads a sequence of moves and reports the resulting state, emitting the state after each move as a short running trace before the final answer. 
Each state has two parts: a \emph{visible coordinate} that can be read directly from the moves, and a hidden coordinate that depends on the \emph{order} in which the moves were applied and that the visible coordinate alone does not determine. 
We write the task as a walk on a finite group, which makes the hidden coordinate precise and gives us ground truth for every intervention.

Formally, a prompt is a sequence of moves \(m_1,\dots,m_T\) drawn from a finite group \(G\) acting on itself by left multiplication. The states satisfy
\[
s_0=e,\qquad s_t=m_t\,s_{t-1},\quad t=1,\dots,T.
\]
The model reports \(s_T\); running-state supervision additionally asks for each intermediate \(s_t\). 
Each state has a \emph{visible coordinate} \(v(s)\) and a \emph{hidden bit} \(h(s)\):
\[
s \longleftrightarrow \bigl(v(s),h(s)\bigr),\qquad
v(s)\in V=\mathbb{Z}_2\times\mathbb{Z}_2,\quad h(s)\in\mathbb{Z}_2.
\] 
The visible coordinate is a running tally of the moves that ignores their order, so it can be read straight off the sequence; the hidden bit instead depends on the \emph{order} in which the moves were applied, so it can only be obtained by tracking the sequence as it is consumed. 
We use the two non-commutative groups of order eight, the quaternion group \(Q_8\) and the dihedral group \(D_8\). 
Both have the same visible quotient \(V\), the Klein four-group, but differ in how their multiplication updates the hidden coordinate. 
The group-theoretic construction is deferred to \Cref{app:groups}; \Cref{fig:prompt-example} gives a worked example prompt.

\paragraph{Why this setting.} 
Prior work has studied representations of cyclic states, such as days and months \cite{engels2024linear,kantamneni2025trig,feucht2026calculator}, and how chain-of-thought can emulate finite automata \cite{zhang2025fsa}. 
Here, changing the hidden bit while holding the visible coordinate fixed gives a counterfactual state whose continuation is determined by the next move. 
This lets us test whether the model updates the edited state according to the task rule.


Unless otherwise stated, we analyze a \texttt{Qwen2.5-Coder-7B} model trained with full running-state supervision; auxiliary supervision arms are reported in \Cref{app:training-arms}.

\section{Counterfactual register editing}
\label{sec:assay}

To test whether the intermediate token carries an executable state, we edit a low-rank \emph{register} at the current-state token and ask whether the model recomputes its next state from the edited value. 
The method has three parts: identify a low-rank state register, edit it to a counterfactual value, and score whether the next-step computation follows the edited state (\Cref{fig:hero} and \Cref{fig:taskOrder}).


\paragraph{Intervention.} Let \(z_{\ell,i}(x) \in \mathbb{R}^d\) denote the residual stream at layer \(\ell\) and token position \(i\) on prompt \(x\), and let \(i^\star\) be the position of the current-state token. 
We estimate the register subspace \(U \subseteq \mathbb{R}^d\) at \((\ell, i^\star)\) from the class-conditional means of the residual
stream: for each visible value we take the difference of means between the two hidden values, and \(U\) is the span of the leading \(r=16\) such directions at the validation-selected layer~12 edit site.
The subspace \(U\) and the class means \(\mu(s)\) are estimated on a calibration split disjoint from all items used to report \CUS\ and register agreement, and the layer and site are locked on a validation split before the causal test.
Writing \(\Pi_U\) for the orthogonal projector onto \(U\) and \(\mu(s)\) for the mean residual stream in state \(s\), the counterfactual state flips only the hidden coordinate, and the edit adds the projected class-conditional mean difference:
\[
\tilde{s} = (v,1-h),\quad
\tilde{z}_{\ell,i^\star} = z_{\ell,i^\star} + \Pi_U\!\bigl(\mu(\tilde{s})-\mu(s)\bigr).
\]
The decoded hidden coordinate reads \(h(\tilde{s})\) while the visible coordinate and the printed token are unchanged. 
The \emph{random-subspace} and \emph{orthogonal-complement} controls apply the same-rank edit in a random subspace or within \(U^{\perp}\); the \emph{full-residual} control replaces the entire token, \(\tilde{z}_{\ell,i^\star} = \mu(\tilde{s})\). 
To check that the hidden coordinate exists at all we also fit a linear readout of it from the same token and report its held-out accuracy, which distinguishes a decodable variable from a causally used one \cite{geiger2024das,turner2023actadd}.


We provide the correct sequence of intermediate states up to the intervention, then intervene on the residual stream at the current hidden-bit token, leaving the literal token unchanged, and measure the model's prediction of the next hidden bit.
\Cref{app:freerun} separately tests whether the register is still used when the model generates the trace itself.


\paragraph{Evaluation.}\label{sec:cus} Let \(\hnext\) be the model's predicted hidden coordinate of the next state under the edit, given the upcoming move \(m\). 
If the model executes its learned update on whatever state is written into the register, then \(\hnext\) should equal \(h(\msp)\), the hidden bit of the move applied to the \emph{edited} state, rather than \(h(\ms)\), the move applied to the \emph{original} state. On the discriminating subset \(\mathcal{D} = \{\, h(\msp) \neq h(\ms) \,\}\) we define \emph{register agreement} and \emph{Counterfactual Update Selectivity},
\[
R \;=\; \mathbb{P}\bigl(\hnext = h(\msp)\bigr),
\qquad
\CUS \;=\; \mathbb{P}\bigl(\hnext = h(\msp)\bigr) - \mathbb{P}\bigl(\hnext = h(\ms)\bigr).
\]
\(R\) measures agreement with the counterfactual next state while \CUS\ measures preference over the original next state.

Two controls separate supplying a state from supplying an answer.
The \emph{move-swap} selectivity varies the move on a fixed edited state, evaluated on items where \(h(\msp)\neq h(m'\!\cdot\!\tilde{s})\):
\[
T = \mathbb{P}\bigl(\hnext=h(\msp)\bigr)
   - \mathbb{P}\bigl(\hnext=h(m'\!\cdot\!\tilde{s})\bigr).
\]
The \emph{conflicting-continuation} selectivity injects a real occurrence \(\tilde{s}_{\mathrm{src}}\) of the counterfactual state whose own continuation \(h_{\mathrm{src}}\) conflicts with the move-applied target, \(h(\msp)\neq h_{\mathrm{src}}\):
\[
L = \mathbb{P}\bigl(\hnext=h(\msp)\bigr)
   - \mathbb{P}\bigl(\hnext=h_{\mathrm{src}}\bigr).
\]

\paragraph{Localization.} 
We localize the update by measuring how ablation, matched-control, and restoration interventions affect \CUS\ \cite{conmy2023acdc,syed2023attribution,meng2022rome,todd2024fv}. 
Further, we test a fixed set of attention edges and contiguous groups of low rank components on the path to the next-state readout.
An ablation masks a candidate component \(c\) during the forward pass and recomputes \(\CUS_c\); the \emph{relative reduction} is \(\kappa_c = (\CUS_{\varnothing} - \CUS_c)/\CUS_{\varnothing}\), where \(\CUS_{\varnothing}\) is the unintervened baseline. 
A matched control routes \(c\) from the visible coordinate rather than the hidden coordinate, and a restoration recovers \(c\)'s contribution from the clean edited run. 
\Cref{tab:q8loc} lists the candidate components and \Cref{app:metrics} provides the predefined localization criteria. 
\section{Trace supervision creates an executable state register}
\label{sec:results}

\ifshowaddedmaterial
We organize the evidence as an edit-site check and three causal tests, each ruling
out a specific alternative explanation:
\begin{enumerate}[leftmargin=1.6em,itemsep=2pt,topsep=3pt]
\item \textbf{Edit-site check}: is the literal hidden-bit token even readable? It is
  the literal trace token, so it decodes in every arm and does not distinguish them: the causal
  evidence begins with \emph{use}.
\item \textbf{Use}: does editing that coordinate change what the model computes next?
  A rank-16 edit redirects the next state (\mbox{\Cref{sec:use}}).
\item \textbf{Move-specific}: does the model apply the \emph{actual} upcoming move to
  the edited state, rather than emit a fixed answer? Swapping the move swaps the next state in step
  (\mbox{\Cref{sec:use}}).
\item \textbf{Conflicting continuation}: does the model recompute its next state from the
  edit, or copy a pasted-in continuation? When the injected state carries a conflicting continuation,
  the model follows the current move and not that continuation
  (\mbox{\Cref{sec:use}}).
\end{enumerate}
\Cref{fig:receipt} summarizes these matched tests, and \Cref{tab:summary} reports their values.

The current state is perfectly decodable as it is written into the trace.
On 600 held-out items, late-layer decoding accuracy for the next-state coordinate exceeds 0.96 in both $Q_8$ and $D_8$ as exhibited in \Cref{fig:decodability}.

\subsection{Editing the register redirects the computation}
\label{sec:use}

On held-out items, a rank-16 same-position edit makes the running-state model apply the upcoming
move to the counterfactual state: next-state agreement is 0.80 in \(Q_8\) and 0.91 in
\(D_8\), versus about 0.02 for equal-rank random and orthogonal-complement edits. Full-residual
replacement is no stronger, so the compact subspace captures the causal effect available at the
site. Because the edit is made at the current-state token, before the next-state computation and
away from the answer, it redirects an intermediate computation rather than simply pushing the
output.\

\begin{figure}[tb]
\centering
\includegraphics[width=\linewidth]{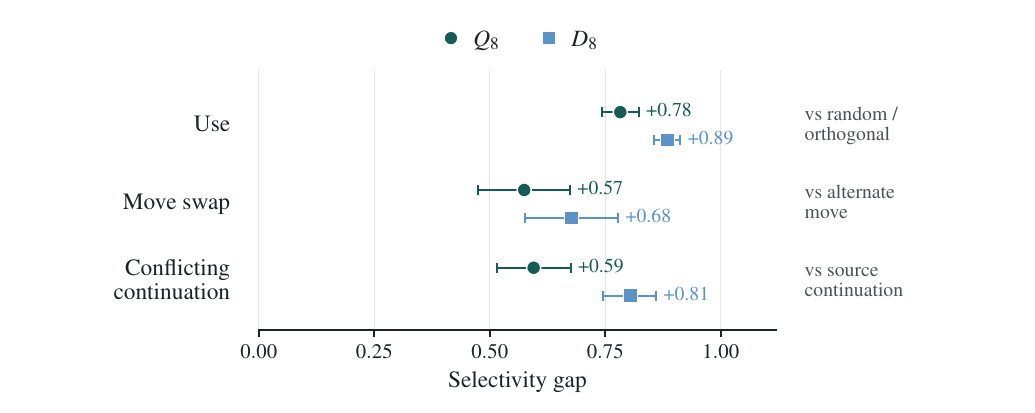}
\caption{\textbf{Summary of the causal chain.} The three rows report complementary selectivity
gaps testing whether the edited coordinate is used as an executable state rather than merely being
readable. Each row uses the matched alternative shown at right, so their magnitudes are not directly comparable. Teal circles denote \(Q_8\), blue squares denote \(D_8\), and whiskers are
95 percent bootstrap intervals on held-out items; all exclude zero.
}
\label{fig:receipt}
\end{figure}

Two controls separate supplying a state from supplying an answer. In the move-swap
control, holding the edited state fixed and changing the upcoming move changes the next state in
the corresponding way: the move-specific next state exceeds the alternative by 0.57 in
\(Q_8\) and 0.68 in \(D_8\). In the conflicting-continuation control, we inject a single real
occurrence of the counterfactual state whose own continuation conflicts with the move-applied
target. The model follows the current move, \(\msp\), on 0.80 of items and the injected
continuation on only 0.20, a selectivity of +0.595; the same control on \(D_8\) gives +0.805, so the edit supplies a state and not an answer in both groups. A single
injected occurrence is as effective as the constructed counterfactual, so the effect is not an
artifact of averaging. Together these controls show the model runs its learned update on whatever
state is written into the register, rather than emitting a fixed answer.

In \texttt{Mistral-7B}, register agreement reaches 0.93 in $Q_8$ and 0.94 in $D_8$. 
Move-swap and conflicting-continuation selectivities range from 0.84 to 0.89, replicating the operation-dependent use of the edited state.


We also edit traces generated by the model, retaining only those that are correct through the edited token in order to test whether the redirection persists.
Indeed, we find persistence through later states and the final answer: in \(D_8\), the edited branch is followed on about 0.86 of items at each of the next three states, whereas \(Q_8\) falls below the one-step criterion but retains above-chance multi-step persistence and positive move-specific and conflicting-continuation controls.


\subsection{Localization}
\label{sec:localization}
The register edit is low-rank in both groups, but the route that updates it differs.
The counterfactual effect saturates by rank~1 in \(Q_8\) and rank~2 in \(D_8\), far below the
3584-dimensional residual stream. In \(Q_8\), ablating a single layer-22 attention edge from
the current-state hidden coordinate to the next-state readout removes 94\% of \CUS; matched
controls preserve the effect, restoration recovers it, and the result replicates on an independent
split. In \(D_8\), no single component removes more than 24\%. A frozen four-edge route
removes 68.7\% of \CUS across ten reported splits, eight of which are held-out confirmatory,
supporting a partial distributed localization. Thus, a compact register need not have a similarly
concentrated update route. Appendix~\ref{app:results} reports complete results; Appendices~\ref{app:second-family}--\ref{app:freerun} cover replication and generated traces, and Appendix~\ref{app:localize} details localization.

\section{Comparisons with Modern Steering Techniques}

The preceding experiments identify a causal register whose value can be edited to redirect subsequent computation.
We now ask whether existing activation-steering methods act through the same mechanism.

\subsection{Function Vector Steering}
We use Function Vectors (FVs) \cite{todd2024fv} to compare activation steering with counterfactual register editing. 
We first test whether FV steering writes into the register identified in the state-tracking model.
We then examine whether effective steering in lexical tasks recreates the model’s natural target activation.

To illustrate the construction, consider English-to-Spanish translation with ten input-output demonstrations followed by a query.
For each query, we construct a correct prompt and a shuffled prompt. 
In the correct prompt, each input is paired with its correct output. 
In the shuffled prompt, the same inputs, the same set of outputs, and the same query are retained, but the outputs are reassigned across demonstrations using a fixed permutation. 
For example,
\emph{dog}~\(\rightarrow\)~\emph{perro} and
\emph{cat}~\(\rightarrow\)~\emph{gato} might become
\emph{dog}~\(\rightarrow\)~\emph{gato} and
\emph{cat}~\(\rightarrow\)~\emph{perro}.
This breaks the task demonstrated by the examples while preserving the prompt content and final answer position, giving a controlled prompt on which steering can restore the target behavior.

The method identifies attention heads whose outputs most improve the target behavior when inserted into shuffled prompts. 
It averages the selected head outputs at the final prompt token over 45 correct prompts and projects the result back into the model's hidden representation. 
Following the original work, we call the resulting direction the FV. 
The details of how the FV was constructed for the $Q_8$ and $D_8$ task are in appendix \ref{app:fv-steering-find}. 

\subsection{Function Vector steering does not directly write the identified registers}

FV steering is effective by the usual behavioral criterion. 
At the intervention scale selected on the training set, it produces the target one-step next state on 65.6\% of training examples and 64.3\% of held-out \(Q_8\) examples. 
This approaches the efficacy of counterfactual register editing, which achieves one-step register agreement of 80\% and free-running final-answer agreement of 72\%. 

Judged only by their outputs, the interventions could therefore appear to change the same internal state.
Their immediate effects on the residual stream are different. 
On held-out examples, the applied update is 2.19\texttimes{} larger than the matched register write and, when added to the same clean residual, produces a residual 1.68\texttimes{} larger. 
More importantly, the two updates have median cosine similarity 0.254, and only 5.7\%--13.4\% of the FV energy lies in the rank-16 register subspace. 
Thus, comparable behavioral success does not imply a comparable local edit. 
Counterfactual register editing writes within the identified causal subspace, whereas the successful FV intervention is a larger, mostly out-of-subspace update. 

These measurements do not rule out another register, a distributed representation, or the possibility that later layers transform the FV update into the identified register.
They demonstrate that successful FV steering is not itself a direct write into the register isolated previously. 
This motivates a broader question: when a steering vector changes behavior in standard tasks, does it recreate an activation that the model naturally uses, or control the answer through a different and unusually large perturbation? 
We test these alternatives next.
\section{Testing the correspondence hypothesis}
\label{sec:steering-assay}

The register experiment gives a clear positive example: after we write a counterfactual value,
the model applies a known transition to that value. We now ask what can be concluded when an
additive steering vector changes an answer. \Cref{tab:causal-roles} summarizes the evidence needed
for three increasingly strong interpretations of an activation intervention.

\subsection{Comparing modern steering techniques with the model's own activations}
We continue our studies using Function Vector Steering. We use two models, and for each model, we choose a task, layer, and steering strength on separate calibration data
before running the comparison below. The selected settings are \textbf{present-to-past conversion} in Qwen
at block 7 with coefficient 15, and \textbf{English-to-Spanish translation} in Llama at block 8 with
coefficient 10. Each evaluation uses 64 new prompts that do not appear in the construction or
calibration data. Because these settings were chosen for a strong steering effect, the experiment
tests how effective steering works rather than how often steering succeeds across tasks or layers.

For evaluation prompt \(i\) and block \(\ell\), let
\(h_{i,\ell}^{\mathrm{shuf}}\) and \(h_{i,\ell}^{\mathrm{corr}}\) denote the hidden vector at the
final prompt token after block \(\ell\) for the shuffled and correct prompts, respectively. Let
\(\ell_\star\) be the selected intervention block, and let \(v\) be the calibrated Function Vector.
All interventions below begin from the shuffled prompt; the corresponding correct prompt supplies
hidden vectors for comparison.

The register experiment established a strong form of local sufficiency: writing one
counterfactual value at one token was enough for a known later operation to act on the edited
value. The lexical tasks below do not provide an explicit state variable with a known update rule,
so we cannot repeat that full semantic test. We instead test a weaker consequence of the same
interpretation: whether successful steering can be explained as a one-token, one-block write of
the hidden vector produced by the correct prompt. We compare five interventions:

\begin{enumerate}[leftmargin=1.4em,itemsep=2pt,topsep=2pt]

\item
\textbf{Repeated correct-prompt patch:} At the selected block and every subsequent block \(\ell\), replace the final-token hidden vector with \(h_{i,\ell}^{\mathrm{corr}}\), leaving earlier token positions unchanged. This provides a reference effect under sustained patching.

\item
\textbf{One-block correct-prompt patch:} Replace \(h_{i,\ell_\star}^{\mathrm{shuf}}\) with \(h_{i,\ell_\star}^{\mathrm{corr}}\) once and let the remaining blocks proceed normally. This tests whether the correct-prompt value is locally sufficient to redirect computation.

\item
\textbf{Mean steering:} Add the calibrated Function Vector \(v\) once at \(\ell_\star\), measuring the behavioral effect of the standard steering intervention.

\item
\textbf{Natural-scale mean steering:} Add \(v\) after rescaling it to \(\lVert h_{i,\ell_\star}^{\mathrm{corr}}-h_{i,\ell_\star}^{\mathrm{shuf}}\rVert_2\). This tests whether steering remains effective at the scale of the natural correct-versus-shuffled activation change.

\item
\textbf{Same-norm comparison directions:} Add the prompt-specific correct-minus-shuffled direction, a held-out average of that direction, another task's Function Vector, or a random direction, each rescaled to \(\lVert v\rVert_2\). These controls test whether the effect is specific to the learned Function Vector.

\end{enumerate}

The one-block patch tests whether the model’s target activation is sufficient at the intervention site. 
Natural-scale steering tests whether the steering direction remains effective at the magnitude of the matched activation change. Together, these interventions test the two predictions of CH.


\subsection{Scoring}

For each prompt \(i\), let \(\Delta_i(c)\) be the change in the target-minus-alternative log-probability margin under intervention \(c\), relative to the unedited shuffled prompt, and let \(G_i\) be the corresponding gain from the repeated correct-prompt patch. Across \(n\) evaluation prompts, the normalized effect is
\[
E(c)=\frac{\frac{1}{n}\sum_{i=1}^{n}\Delta_i(c)}
           {\frac{1}{n}\sum_{i=1}^{n}G_i}.
\]
Thus, \(E(c)=0\) denotes no average improvement and \(E(c)=1\) denotes the same average margin gain as the reference. Values above one are possible because the reference is not an upper bound. This score compares behavioral effects, not similarity between hidden vectors. We report paired 95\% bootstrap intervals from 2,000 resamples; full definitions and normalization details are in \Cref{app:steering-scoring}.

\section{Steering does not reproduce the target activation}
\label{sec:steering-results}

The register experiments show that a single local activation edit can be sufficient to redirect subsequent computation: once the state is changed, the model continues from the edited value. We now ask whether the activation targeted by Function Vector steering has the same kind of local sufficiency. In particular, does the model’s own target-task activation work when transplanted once at that site, and does effective steering operate at the scale of this natural change?

\subsection{The model's target activation is not sufficient}

The all-layer target patch selects the target answer on all 64 prompts and increases the
target-minus-alternative margin by 4.41 nats in Qwen and 2.64 nats in Llama, establishing the
reference effect. Patching the matched target activation only at the tested layer produces almost
none of this gain: its normalized effect is 0.014 [0.005, 0.023] in Qwen and
\textminus{}0.015 [\textminus{}0.041, 0.005] in Llama (\Cref{fig:steering-comparison}a).

The natural-scale test also fails. The calibrated Function Vectors have norms 210.80 in Qwen and
20.22 in Llama, 77 and 26 times the median source-to-target changes of 2.74 and 0.78. Rescaling the
vectors to those natural magnitudes reduces their normalized effects to 0.007 and 0.017.
At the calibrated scale, they produce 0.544 and
0.824 of the all-layer patching gain. 

\begin{figure}[tb]
\centering
\includegraphics[width=0.52\linewidth]{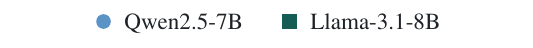}\\[-2pt]
\begin{subfigure}[t]{0.49\linewidth}
\centering
\includegraphics[width=\linewidth]{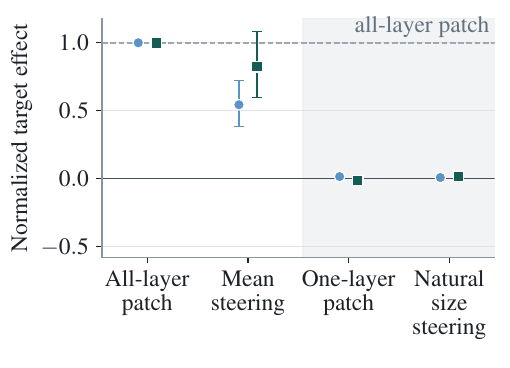}
\caption{Target activation versus mean steering}
\end{subfigure}\hfill
\begin{subfigure}[t]{0.49\linewidth}
\centering
\includegraphics[width=\linewidth]{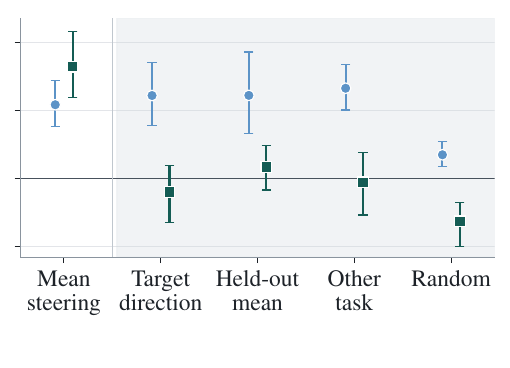}
\caption{Specificity differs across models}
\end{subfigure}
\caption{\textbf{Effective steering does not reproduce the model's target activation at one
layer.} (a) Patching target activations through all later layers is the normalized reference.
Patching the target activation at only the tested layer and applying the steering vector at the
scale of the natural activation change both have effects near zero. The conventional, much larger
steering vector changes the answer in both models. (b) Llama responds to the task's steering
direction but not to equally large comparison directions. Qwen responds to several structured
directions, although not as strongly to a random direction. Points show ratios of paired mean
margin changes; bars are paired 95\% bootstrap intervals over 64 new prompts per model.}
\label{fig:steering-comparison}
\end{figure}


Neither prediction of CH holds in the two tested settings: the matched target activation and natural-scale steering have little effect, whereas a much larger steering vector changes the answer. 
Successful steering therefore need not recreate the model’s natural target activation at the tested layer.

\subsection{The two models differ in steering specificity}

Because the calibrated steering vectors are much larger than the models' natural activation
changes, we use equal-norm controls to test whether their effects depend on direction
(\Cref{fig:steering-comparison}b). In Llama, the target task's Function Vector has a normalized
effect of 0.824, while the four comparison directions range from \textminus{}0.316 to 0.085;
among the directions tested, the effect is therefore selective. Qwen behaves differently: three
structured comparison directions have effects between 0.612 and 0.665, comparable to or
larger than the target Function Vector's 0.544, although a random direction is weaker at
0.175. Thus, scale alone does not determine the effect, but specificity is model dependent.
Even in Llama, specificity strengthens the mechanistic interpretation without showing that
steering recreates a natural target activation or writes a state-like value.

\subsection{Additional tests narrow the interpretation}

These tests do not identify what the large vector does, but they provide evidence against two
additional simple explanations. Across two models and four lexical tasks, we add the vector at its
usual layer and subtract it at progressively later layers. Early subtraction eliminates the
task-specific effect; the effect survives only when subtraction occurs late, at depths where
matched comparison directions also begin to affect the answer. We therefore find no evidence that
the early steering pulse is quickly converted into a self-sustaining task state. When the
underlying task is held fixed but the output labels are remapped, the unedited models follow the
new mapping, whereas the steering effect does not reverse with it. Part of the effect remains tied
to the same literal label, arguing against an abstract ``correct answer'' signal. Full results appear in \Cref{app:steering-diagnostics}.



\section{Discussion}
\label{sec:discussion}

The controlled state-tracking experiments provide direct evidence for a causal register in the
residual stream. Writing a counterfactual value \(\tilde{s}\) into the low-rank component at the
current-state token causes the model to apply the actual upcoming move \(m\) and predict
\(m(\tilde{s})\). The consequence of the same write changes with the upcoming move, while a
conflicting continuation associated with the injected state does not determine the prediction.
The counterfactual branch also persists during free-running generation, cleanly in \(D_8\) and
more partially in \(Q_8\), and the core result replicates in a second model family. Together,
these tests establish a causal register: the value written at the trace token is read by later
computation and updated according to the task rule.
The causal effect saturates at rank one in \(Q_8\) and rank two in \(D_8\), although the update
route is concentrated in \(Q_8\) and more distributed in \(D_8\). Thus, the dimensionality of a
stored variable and the concentration of the computation that uses it are separate properties.

The Function Vector interventions present a different picture. Replacing the final-token residual
once with the exact residual from the matched correct prompt has almost no effect, and a Function
Vector rescaled to the natural correct-versus-shuffled change is similarly ineffective. At its
calibrated scale, the vector is effective, but its norm is 77 times the median natural
change in Qwen and 26 times that change in Llama. Successful steering at these sites is
therefore not well described as restoring the residual that the model naturally reaches on the
target task. This is the narrow, measured sense in which the effective intervention is
\emph{unnatural}: its immediate displacement at the tested site is tens of times larger than the
matched natural change.

Subtracting the vector at an early downstream layer eliminates the task-specific effect; much of the effect survives only when subtraction is delayed until late in the network. 
Under this test, we therefore find no evidence that the initial perturbation is quickly converted into an independent, self-sustaining task state. 
Answer-remapping and equal-norm controls further show that the vector is not simply an abstract ``correct answer'' signal and that scale alone does not determine the effect. 

\paragraph{Limitations.}
The register experiments use explicit traces in controlled tasks, while the steering experiments examine one selected effective task and layer per model, with intervention at the final prompt token. 
The one-block patch therefore tests restoration only at that site, leaving open states represented elsewhere, distributed across
several activations, or constructed downstream.
Likewise, \emph{unnatural} refers only to the displacement relative to the matched natural change, and cancellation establishes continued dependence on that displacement without identifying how later layers use it. 

A register write is identified by what later computation does with the edited value: after one local edit, the model continues from that value under the known update rule. 
Behavioral success alone does not establish such a write.


\begin{ack}
Benjamin Shih acknowledges support from Perpetual Labs, including computational resources used in this work.
\end{ack}

\newpage
\appendix
\crefalias{section}{appendix}
\crefalias{subsection}{appendix}
\crefalias{subsubsection}{appendix}

\begingroup
\showaddedmaterialfalse
\color{black}

\section{Task and experimental setup}
\label{app:setup}
\suppressfloats[t]

The register experiments use a known transition rule to distinguish changing a state from changing an answer. This appendix collects the state-space construction, prompt and data specifications, and model-training settings.

\FloatBarrier
\subsection{State space and token vocabulary}

\label{app:groups}

The two move systems are the quaternion group \(Q_8\) and the dihedral group \(D_8\),
the two non-commutative groups of order eight. We use them because each is a central
extension of the Klein four-group \(V = \mathbb{Z}/2 \times \mathbb{Z}/2\) by
\(\mathbb{Z}/2\):
\[
1 \longrightarrow \mathbb{Z}/2 \longrightarrow G \longrightarrow V \longrightarrow 1,
\qquad G \in \{Q_8, D_8\}.
\]
This gives every state \(g \in G\) a two-part description. The image of \(g\) in the
quotient \(V\) is the \emph{visible coordinate}: a four-way value that a model can read
directly from the move sequence, because the quotient multiplication is commutative.
The remaining \(\mathbb{Z}/2\) factor is the \emph{hidden coordinate}: a single bit
fixed by the group's two-cocycle, which records the order in which moves were applied.
\(Q_8\) and \(D_8\) are distinguished precisely by their cocycle, so a hidden-coordinate
result that holds for both is a property of the supervision and not of one cocycle.

\paragraph{Order sensitivity.} The hidden bit is what makes the task order-sensitive.
For two moves \(a, b\) whose visible parts commute, the products \(ab\) and \(ba\) share
a visible coordinate but differ in the hidden coordinate whenever the cocycle is
nontrivial on that pair. A model that tracks only the visible coordinate, or only the
move multiset, cannot distinguish \(ab\) from \(ba\); a model that carries the hidden
coordinate can. The held-out loop tests are sequences
that return to the identity in the quotient but differ in the hidden bit, so they pass
only if the order-sensitive coordinate was learned.

\paragraph{Labels.} In figures and tables a state is written as \(s = (v, h)\): a visible
value \(v \in \{00, 01, 10, 11\}\) and a hidden bit \(h \in \{0, 1\}\), written \texttt{v|h}
(for example \texttt{11|1}). The model itself reads and writes a small fixed vocabulary of single
tokens, a \emph{region} letter for the visible value, a \emph{phase} letter for the hidden bit,
and a move letter for each generator, chosen so that every token is a single subword and the
current-state and next-state positions are fixed across prompts (\Cref{app:datasets}).
\Cref{tab:vocab} records the exact correspondence we use to relabel that raw vocabulary as
coordinates throughout the paper; these letters are the \texttt{J}, \texttt{M}, \texttt{F},
\texttt{C} (and \texttt{P}, \texttt{R}) that appear in the model's own input and output. The
labels are a presentation device for the eight states; the operational content is the quotient
and the central bit defined above.

\begin{table}[!htbp]
\centering
\small
\caption{\textbf{Token vocabulary and its coordinate relabeling.} The four region letters and two
phase letters are the eight raw state tokens (\texttt{J}, \texttt{M}, \texttt{F}, \texttt{C},
\(\dots\)) the model reads and writes; the move letters \texttt{Q}, \texttt{T}, \texttt{W},
\texttt{Z} are the four generators \(a, b, a^{-1}, b^{-1}\). The inverse moves \texttt{W}, \texttt{Z}
project to the same visible bit flips as \texttt{Q}, \texttt{T} but differ in their effect on the hidden bit,
so the quotient walk is identical across \(Q_8\) and \(D_8\) while the hidden bit is not.}
\label{tab:vocab}
\begin{tabular}{@{}cc@{\hspace{2.5em}}cc@{\hspace{2.5em}}cc@{}}
\toprule
\multicolumn{2}{c}{\textbf{Region} \(v\) (visible)} & \multicolumn{2}{c}{\textbf{Phase} \(h\) (hidden)} & \multicolumn{2}{c}{\textbf{Move} (generator)}\\
token & coord.\ & token & bit & token & action\\
\midrule
\texttt{C} & \texttt{00} & \texttt{P} & \(0\) & \texttt{Q} & flip 1st bit \((a)\)\\
\texttt{F} & \texttt{10} & \texttt{R} & \(1\) & \texttt{T} & flip 2nd bit \((b)\)\\
\texttt{J} & \texttt{01} &           &       & \texttt{W} & \(a^{-1}\)\\
\texttt{M} & \texttt{11} &           &       & \texttt{Z} & \(b^{-1}\)\\
\bottomrule
\end{tabular}
\end{table}

\Cref{fig:filmstrip} shows the cocycle difference on a single worked example: the same move
sequence reaches the same visible cells in \(Q_8\) and \(D_8\), but the hidden bit diverges in the
middle of the walk.

\begin{figure}[!htbp]
\centering
\makebox[\linewidth][l]{\hspace*{-0.013\linewidth}\includegraphics[width=\linewidth]{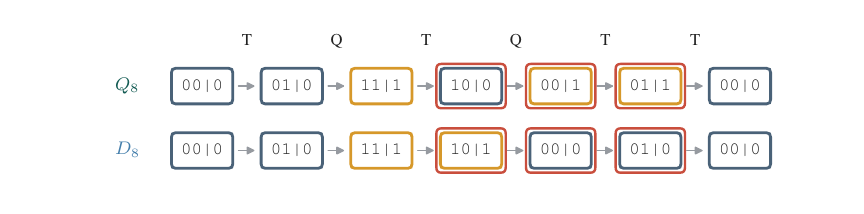}}
\caption{\textbf{Same moves, same visible path, different hidden bit.} The move sequence
\texttt{T Q T Q T T} from \(\texttt{00}\,|\,0\) produces the same sequence of visible coordinates in
\(Q_8\) and \(D_8\) (the quotient walk is identical), but the hidden bit \(h\) (amber \(h{=}1\),
steel \(h{=}0\)) diverges where boxed. The cocycle that separates the two groups, on one worked
example.}
\label{fig:filmstrip}
\end{figure}

\Cref{fig:cubetwist} makes the same difference structural: with the hidden bit drawn as a vertical
axis, the move \(T\) stays within an \(h\) level in \(D_8\) (order 2) but threads both levels in
\(Q_8\) (order 4); the quotient walk on the visible cells is identical in both groups, and only the
hidden-bit update differs.

\begin{figure}[!htbp]
\centering
\includegraphics[width=0.92\linewidth]{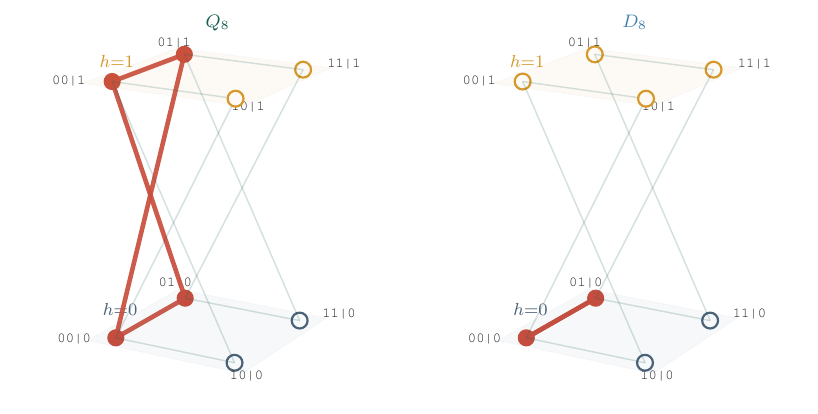}
\caption{\textbf{How the \(Q_8\) and \(D_8\) cocycles differ.} The eight states
are the corners of a cube whose vertical axis is the hidden bit \(h\); the four visible cells sit in
each level. The move \(Q\) (teal) is identical in both groups. The move \(T\) (coral orbit from
\(\texttt{00}\,|\,0\)) stays within one level in \(D_8\), so it has order 2; in \(Q_8\) it flips \(h\)
and threads both levels, so it has order 4. The quotient walk on the visible cells is the same in
both groups; only the hidden-bit update differs, and that extra turn through the hidden bit is the
cocycle that separates them.}
\label{fig:cubetwist}
\end{figure}

\FloatBarrier
\subsection{Prompts, datasets, and evaluation splits}

\label{app:datasets}

\paragraph{Prompt format.} Each item presents an initial state and a sequence of moves,
and asks for the resulting state. In the running-state condition the target sequence
emits the intermediate state after each move before the final answer; in the final-answer
condition the target is the final answer only. Moves and states are written with a small
fixed vocabulary so that the current-state token and the next-state readout occur at
known positions, which is what makes the register edit and the edge interventions
positionally well defined; the worked example in \Cref{fig:prompt-example} shows
the format.

\paragraph{Sequences and splits.} Items are generated by sampling move sequences over
the group and computing the induced state trajectory. The supervised fine-tuning corpus for
each group (\(Q_8\), \(D_8\)) holds 11,346 training items, 3,782 distinct move
sequences of length 1--6, phase-balanced within each length and move-multiset bucket,
each rendered under three instruction paraphrases, together with 856 held-out validation
items (a single held-out template on disjoint sequences). We hold out a disjoint split of
items for all reported measurements, and the localization analysis uses a second,
independently sampled split to test replication. Counterfactual items for the register
edit are constructed by pairing each held-out current state with its same-visible,
opposite-hidden partner; conflicting-continuation items additionally require a real occurrence
of the partner state whose continuation disagrees with the move-applied target, so the
two outcomes are discriminating by construction.

\paragraph{Discriminating items.} The selectivity metrics are evaluated only on items
where the two targets being compared differ, so that a difference in outcome is
attributable to the intervention. Each reported selectivity metric uses \(n = 400\) held-out
discriminating items; the layer-resolved decodability of \Cref{fig:decodability} uses
\(n = 600\) held-out items per layer. Move sequences are sampled to balance the eight states
and the four generators; the full dataset, with per-state and per-move counts, is released
with the code.

\paragraph{Prompt templates.} The model reads the raw single-token vocabulary of
\Cref{tab:vocab}. Held-out items share the prefix
\texttt{\small Walk. Start region \{region\} phase \{phase\}. Moves \{moves\}}
and end with \texttt{\small Final region then phase:} (final-answer arm) or
\texttt{\small State trace, then final region and phase:} (running-state arm, whose target
emits each intermediate \texttt{\small region phase} pair before the final answer). Training
uses three paraphrases of this instruction; the held-out split uses the template above, so
the evaluation wording is disjoint from training.

\begin{figure}[!htbp]
\centering
\begin{subfigure}[t]{0.49\linewidth}
\centering
\includegraphics[width=\linewidth]{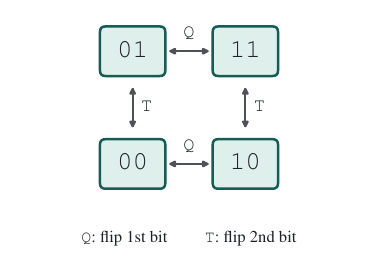}
\caption{Each move flips one visible bit}
\end{subfigure}\hfill
\begin{subfigure}[t]{0.49\linewidth}
\centering
\includegraphics[width=\linewidth]{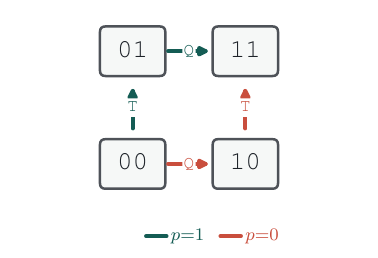}
\caption{Order sets the phase bit}
\end{subfigure}\\[7pt]
\setlength{\fboxsep}{8pt}
\fcolorbox{controlgray}{white}{%
\begin{minipage}{0.93\linewidth}
\small
{\rmfamily\bfseries Prompt (user).}\\[2pt]
\texttt{Walk. Start state 00|0. Moves T Q T Q T T}\\
\texttt{State trace, then final state:}\\[6pt]
{\rmfamily\bfseries Running-state output (model).}\\[2pt]
\texttt{01|\hidphase{0}} $\to$ \texttt{\editsite{11|1}} $\to$ \texttt{10|\hidphase{0}} $\to$ \texttt{00|\hidphase{1}} $\to$ \texttt{01|\hidphase{1}} $\to$ \texttt{00|\hidphase{0}}\\[2pt]
\texttt{Final: 00|0}
\end{minipage}}
\caption{\textbf{The task as a walk on a small state space, with the hidden coordinate and an
edit site highlighted.} A state is a visible coordinate
\(v \in \{\texttt{00},\texttt{01},\texttt{10},\texttt{11}\}\) (one of four cells of the
\(2\times2\) grid) and a hidden bit \(h\), written \(v\,|\,h\). (a)~Each move flips one visible
bit, so \(v\) is a function of the move multiset and easy to read. (b)~The hidden bit is
order-dependent: the same moves in a different order can reach the same cell with the opposite
\(h\), so it must be tracked as the sequence is consumed. An example prompt and the model's
running-state output are shown in these coordinates (the model's literal single-token vocabulary and
its mapping to \(v\,|\,h\) are in \Cref{tab:vocab}). To run the counterfactual edit of
\Cref{sec:assay} we take one current-state token (amber, here \texttt{11|1}), replace a rank-16
subspace of its residual stream so the hidden bit reads the opposite value
(\(\texttt{11|1}\to\texttt{11|0}\)) while the printed token is unchanged, and read the model's next
state: if the register is executable, the next hidden bit follows the move applied to the edited
state, not the original one.}
\label{fig:prompt-example}
\label{fig:taskOrder}
\label{Visualization}
\end{figure}

\FloatBarrier
\subsection{{Use of Large Language Models}}
\label{app:llmusage}
We used large language models for routine coding assistance and for proofreading and formatting the manuscript.

\FloatBarrier
\subsection{Models, training, and intervention sites}

\label{app:models}

\paragraph{Base model.} The primary three arms are built from \texttt{Qwen2.5-Coder-7B}; the
second-family replication (\Cref{app:robustness}) trains the same arms from
\texttt{Mistral-7B-v0.3}. The base arm is the unmodified pretrained model.

\paragraph{Fine-tuning.} The final-answer and running-state arms are produced by low-rank
adaptation of the base model on the task. The two arms use the same move sequences and
the same final answers and differ only in whether the intermediate state is present in
the training target. This matched design is what lets us attribute the causal use of the
register to trace supervision rather than to task exposure or to optimization.

\paragraph{Compute Resources}
All experiments were run on RunPod cloud instances equipped with
NVIDIA A100 or H100 GPUs, each with 80\,GB of GPU memory.
Each experimental run used 1 GPU and required approximately 1 hour. 

\paragraph{Intervention sites.} The register edit acts on the residual stream at the
current-state token at a mid-network layer (layer~12), selected and locked on a validation
split before the causal test; results are reported at this layer and replicated at a
neighboring layer (\Cref{app:rank-robustness}). The localization analysis intervenes on
attention edges into the next-state readout and on low-rank adapter bands across a fixed
range of layers. \Cref{tab:repro} collects the settings needed to reproduce the reported
measurements.

\begin{table}[!htbp]
\centering
\small
\caption{\textbf{Experimental specification} for the reported measurements. Per-state and
per-move balance, exact prompt templates, and dataset item counts are released with the
code and data.}
\label{tab:repro}
\begin{tabularx}{\linewidth}{lX}
\toprule
\textbf{Component} & \textbf{Setting} \\
\midrule
Base model & \texttt{Qwen2.5-Coder-7B} (\(28\) layers, \(d_{\mathrm{model}} = 3584\)) \\
Fine-tuning (LoRA) & rank \(16\), \(\alpha = 32\); target modules
  \texttt{q,k,v,o,gate,up,down\_proj} (all layers) \\
Optimization & learning rate \(10^{-4}\), \(4\) epochs, warmup ratio \(0.05\), max sequence
  length \(256\) \\
Register edit & rank-\(16\) subspace at the current-state token, \texttt{resid\_pre},
  mid-network layer \(12\) (validation-selected edit site; replicated at a neighboring layer) \\
Localization scan & \(38\) candidate components (attention edges into the next-state readout
  and low-rank adapter bands) over layers \(12\)--\(27\); the \(Q_8\) winner is the
  layer-\(22\) edge and the \(D_8\) frozen set is layers \(\{19, 22, 23, 25\}\) \\
Evaluation & \(n = 400\) held-out discriminating items per metric per split; \(Q_8\) localization
  on two independent splits, \(D_8\) route on ten reported splits (eight held-out confirmatory); layer-resolved
  decodability at \(n = 600\) per layer \\
\bottomrule
\end{tabularx}
\end{table}

\paragraph{Reproducibility.} The anonymous review supplement contains the source code, source CSV/JSON
tables, and figure-generation scripts needed to reproduce every
number in the paper.

\FloatBarrier

\section{Causal register results and replication}
\label{app:results}
\suppressfloats[t]

We first collect the evidence that the edited value is used under the supplied transition rule, then test whether that causal chain transfers to a second model family and to generated traces. The second-family and free-running tests, together with the route-null and subset analyses in \Cref{app:d8null,app:d8subset}, were conducted as follow-up analyses after the main results.

Register measurements are reported on held-out items. Selectivity is Counterfactual Update Selectivity
(\CUS, \Cref{sec:assay}); item-level intervals are 95 percent bootstrap intervals over
held-out items, while the \(D_8\) four-edge route interval is computed across the ten
reported splits (eight held-out confirmatory).

\FloatBarrier
\subsection{Supervision controls and the core causal chain}

\label{app:training-arms}

The main analysis focuses on the model trained with full running-state supervision. To separate
the effect of exposing intermediate states from task exposure alone, we also evaluate two matched
controls. The \emph{pretrained} arm is the unmodified base model. The \emph{final-answer-only}
arm is fine-tuned on the same move sequences and final states as the full-supervision arm, but its
target contains only the final state. The \emph{full running-state} arm instead emits the state
after every move before giving the final answer. Thus, the two fine-tuned arms see the same inputs
and final answers and differ only in whether intermediate states appear in the training target.
Exact model and optimization settings are given in \Cref{app:models}.

\paragraph{Readout comparison.}
At the edited current-state token, the hidden coordinate is the literal trace token and therefore
decodes at 1.00 in every arm; edit-site readability does not distinguish the models. The
relevant comparison is the \emph{next}-state coordinate, which must be computed rather than read
from the token. In \(Q_8\), its maximum layerwise probe accuracy is 0.5350 in the pretrained
model and 0.5150 in the final-answer-only model, but rises to 0.9650 in the full
running-state model. In \(D_8\), the pretrained and final-answer-only models already contain a
partial next-state readout, reaching 0.7483 and 0.7567, while the full running-state model
reaches 0.9633. The \(D_8\) result should therefore be interpreted as converting an existing
partial readout into an editable, executable register rather than creating the readout from
nothing \cite{prakash2024entity}.

\begin{figure}[tb]
\centering
\includegraphics[width=\linewidth]{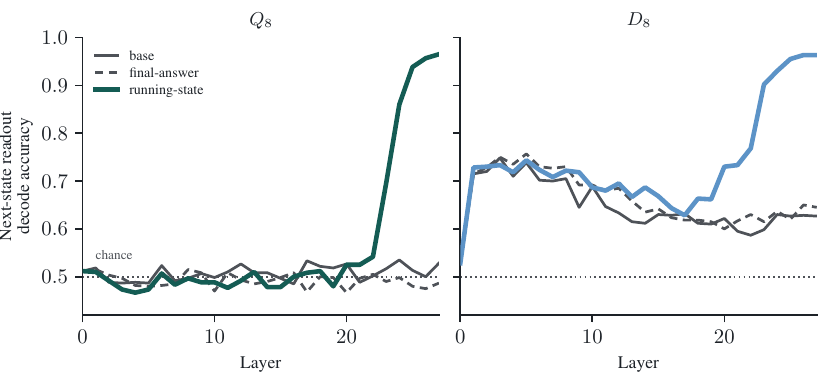}

\caption{\textbf{Decodability of the next-state hidden coordinate by layer.} Linear-probe
accuracy at each layer in the pretrained, final-answer-only, and full running-state models
(chance 0.5). In \(Q_8\), only the full running-state model decodes the coordinate, and only
in the last layers. In \(D_8\), the two auxiliary arms already decode it partially in early
layers, but the full running-state model reaches a higher value in late layers. \(n=600\)
held-out items per point.}
\label{fig:decodability}
\end{figure}

\paragraph{Additional downstream readout.} At the counterfactual-edit site (layer~12, \texttt{resid\_pre}, the current
hidden-bit token), the hidden coordinate is the literal trace token and decodes at 1.000 in
every arm (base, final-answer, and running-state, in both \(Q_8\) and \(D_8\)). Edit-site
readability therefore does not distinguish the arms. The
arm-distinguishing quantity is the \emph{next}-state coordinate the model must compute; a late-layer probe of it (\texttt{phase\_read\_resid\_post}, layer~27) reaches
0.972136 [0.958204, 0.984520] in \(Q_8\) and
0.976266 [0.963608, 0.987342] in \(D_8\) for the running-state model, reported as a downstream readout rather than an edit-site measurement.

\paragraph{Causal comparison.}
Readability alone does not establish that the model uses the coordinate. Under the rank-16
counterfactual register edit, the pretrained and final-answer-only Qwen models remain at chance
on the causal Use test, whereas the full running-state model reaches register agreement 0.80
in \(Q_8\) and 0.91 in \(D_8\). The same three-arm comparison is
repeated for \texttt{Mistral-7B-v0.3}; its exact per-arm Use and selectivity results are reported
in \Cref{app:second-family}.

\paragraph{Move-specific and conflicting-continuation controls.} Move-swap selectivity is +0.5745 in \(Q_8\) and
+0.6768 in \(D_8\). The conflicting-continuation control in \(Q_8\) gives 0.7975 for
following the current move against 0.2025 for following the injected continuation, a
selectivity of +0.5950 with interval [0.5150, 0.6750]; a single injected occurrence
is as effective as the constructed counterfactual. The same control in \(D_8\) gives
0.9025 against 0.0975, a selectivity of +0.8050 with interval [0.7450, 0.8600]
(\(n=400\) held-out items in each antileak row), so the edit supplies a state and not an answer in
both groups.

\begin{table}[!htbp]
\centering
\small
\caption{\textbf{The counterfactual register chain, with a reference level for every row.}
Each row reports a different quantity, named in the row, so the table is read row by row and
not down a single scale. The \emph{null} column gives the value that same quantity takes when
the effect is absent, a chance level or a matched control, and each \(Q_8\)/\(D_8\) entry is
read against it. The edit-site check verifies the hidden bit is readable at the edited token, which is the
literal trace token and so decodes at \(1.00\) in every arm; it is a sanity check, not evidence for
trace supervision, and the causal chain begins with Use. Use is the probability that the next hidden coordinate matches the move applied
to the \emph{edited} state (random and orthogonal-complement edits give \(0.02\)); Move-specific
and Conflicting continuation are selectivity gaps that are \(0\) when the model ignores the edit;
Localize is the fraction of the counterfactual update removed when its attention route is
ablated. The conflicting-continuation control passes in both groups; the \(D_8\) update route is
localized across ten reported splits, eight of them held-out confirmatory (mean \(68.7\%\) reduction, 95\% CI \([66.9, 70.4]\)). The next-state coordinate that the model must \emph{compute}, rather than read off the token, is the arm-distinguishing readout in \Cref{fig:decodability}. Full tables and
intervals are in \Cref{app:results,app:localize}.}
\label{tab:summary}
\begin{tabularx}{\linewidth}{Xccc}
\toprule
\textbf{Step (held-out), with the quantity it reports} & \textbf{null} & \(\boldsymbol{Q_8}\) & \(\boldsymbol{D_8}\) \\
\midrule
Edit-site check: is the literal hidden-bit token readable? (accuracy) & \(1.00\) & \(1.00\) & \(1.00\) \\
Use: does the edit set the next state? \(\mathbb{P}(\hnext = h(\msp))\) & \(0.02\) & \(0.80\) & \(0.91\) \\
Move-specific: does it follow the supplied move? (gap) & \(0\) & \(+0.57\) & \(+0.68\) \\
Conflicting continuation: current move minus injected continuation & \(0\) & \(+0.59\) & \(+0.81\) \\
Localize: fraction of \CUS\ removed by the update route & \(0\) & \(94\%\) & \(68.7\%\) \\
\bottomrule
\end{tabularx}
\end{table}

\paragraph{Supporting layer-stitching and transfer checks.}

In a layer-stitching check, the early and middle layers of the
running-state model drive the later layers of a weaker model, while a weak early
computation is not restored by strong later layers. In a base-coordinate transfer check,
the best base-to-running decoder reaches 0.537 full-state accuracy against a
running self-decoding ceiling of 0.557 and a chance rate of 0.125, with
retention 0.953; a surface-feature control reaches 0.277, above the 0.25
threshold, so this check is supportive but mixed and does not by itself exclude a readout
shortcut. These results corroborate that the supervision builds, rather than only routes,
the computation, and they are not part of the headline causal chain.

\FloatBarrier
\subsection{The core chain in a second model family}
\label{app:robustness}

\label{app:second-family}

We repeated the core counterfactual-register chain on \texttt{Mistral-7B-v0.3}, trained with the
same running-state training procedure and scored on held-out items with a rank-16 register edit, read
out at the layer chosen by the predeclared earliest-readable-current-phase rule. We did not
attempt circuit localization in this model. Because the current-state token is teacher-forced it
is linearly readable in every arm (Read \(=1.00\) throughout), so Read does not distinguish the
arms here; \emph{use} does. Only the running-state arm passes Use and the two selectivity
controls; the base and final-answer arms decode the state but stay near chance on Use and near
zero on the controls (\Cref{tab:second-family}). The executable register chain therefore
replicates in a second model family.

\begin{table}[!htbp]
\centering
\caption{\textbf{Core counterfactual-register chain in \texttt{Mistral-7B-v0.3}.} Exact
per-arm means with 95 percent item-bootstrap intervals. Use and Conflicting continuation use
\(n=2000\); the move-specific column uses the discriminating subset (\(n=1334\) for \(Q_8\),
\(n=967\) for \(D_8\)). Circuit localization was not evaluated in this model.}
\label{tab:second-family}
\scriptsize
\resizebox{\linewidth}{!}{%
\begin{tabular}{llccc}
\toprule
Group & Arm & Use & Move-specific & Conflicting continuation \\
\midrule
\(Q_8\) & running-state & \(0.9265\ [0.9150,\,0.9375]\) & \(0.8471\ [0.8186,\,0.8756]\) & \(0.8530\ [0.8300,\,0.8750]\) \\
\(Q_8\) & base & \(0.5010\ [0.4790,\,0.5225]\) & \(-0.0060\ [-0.0600,\,0.0465]\) & \(0.0020\ [-0.0420,\,0.0450]\) \\
\(Q_8\) & final-answer & \(0.5050\ [0.4830,\,0.5270]\) & \(0.0105\ [-0.0435,\,0.0630]\) & \(0.0100\ [-0.0340,\,0.0540]\) \\
\midrule
\(D_8\) & running-state & \(0.9430\ [0.9325,\,0.9525]\) & \(0.8428\ [0.8077,\,0.8759]\) & \(0.8860\ [0.8650,\,0.9050]\) \\
\(D_8\) & base & \(0.5355\ [0.5130,\,0.5575]\) & \(-0.0527\ [-0.1148,\,0.0093]\) & \(0.0710\ [0.0260,\,0.1150]\) \\
\(D_8\) & final-answer & \(0.4425\ [0.4210,\,0.4645]\) & \(0.0052\ [-0.0569,\,0.0693]\) & \(-0.1150\ [-0.1580,\,-0.0710]\) \\
\bottomrule
\end{tabular}}
\end{table}

\FloatBarrier
\subsection{Free-running register edits and branch persistence}

\label{app:freerun}

This is the free-running test summarized in \Cref{sec:use}. On the
\texttt{Qwen2.5-Coder-7B} running-state models we let the model greedily generate the running
trace, retain only rollouts whose generated assistant prefix matches the correct trace token-for-token
through the current hidden-bit token (the retained fraction is the \emph{coverage}), and apply the
same rank-16 layer-12 \texttt{resid\_pre} same-visible/opposite-hidden edit to the model's own
generated current-state token. We score the one-step counterfactual update, the move-specific and
conflicting-continuation selectivities, and multi-step \emph{persistence}: after the edit we continue the
greedy rollout and ask whether each later hidden state, and the final answer, follows the edited
branch rather than the original branch. The protocol, exact-prefix filter, and a
success threshold of 0.70 for Use were fixed before scoring; \(D_8\) passes the threshold and
\(Q_8\) does not (one-step Use 0.66), and we report both. Intervals are 95\% item-level bootstraps.

\begin{table}[!htbp]
\centering
\caption{\textbf{Free-running counterfactual register editing on \texttt{Qwen2.5-Coder-7B}.} The
running-state model generates its own trace; only rollouts whose generated prefix is exactly correct
through the edited token are scored. The hidden-bit edit redirects the next state and the
redirection persists over the next three states and into the final answer, cleanly in \(D_8\) and
partially in \(Q_8\) (one-step Use \(0.66 < 0.70\)).}
\label{tab:freerun}
\begin{tabular}{lcc}
\toprule
Metric & \(Q_8\) & \(D_8\) \\
\midrule
Coverage (exact-prefix filter) & \(0.464\) & \(0.768\) \\
One-step Use, \(\mathbb{P}(\hnext = h(\msp))\) & \(0.660\)~\([0.597, 0.723]\) & \(0.902\)~\([0.871, 0.934]\) \\
Move-specific selectivity & \(+0.260\)~\([0.099, 0.428]\) & \(+0.637\)~\([0.521, 0.743]\) \\
Conflicting continuation selectivity & \(+0.376\)~\([0.183, 0.570]\) & \(+0.385\)~\([0.180, 0.590]\) \\
Branch persistence, \(k{=}2\) & \(+0.466\)~\([0.343, 0.590]\) & \(+0.722\)~\([0.639, 0.799]\) \\
Branch persistence, \(k{=}3\) & \(+0.403\)~\([0.269, 0.537]\) & \(+0.724\)~\([0.627, 0.813]\) \\
Branch persistence, \(k{=}4\) & \(+0.402\)~\([0.231, 0.556]\) & \(+0.702\)~\([0.590, 0.814]\) \\
Final-answer edited-branch agreement & \(0.723\)~\([0.660, 0.782]\) & \(0.845\)~\([0.808, 0.882]\) \\
\bottomrule
\end{tabular}
\end{table}

\FloatBarrier

\section{Register dimensionality and update localization}
\label{app:localize}
\suppressfloats[t]

A compact register need not be updated by a single concentrated circuit. We bring together the localization protocol, the complete Q8 and D8 results, and the controls that test the identified routes.

\FloatBarrier
\subsection{Candidate scan, decision rules, and route selection}

\label{app:metrics}

The register edit and its control conditions (full residual, random subspace, orthogonal
complement, and the null no-patch and identity conditions), and the metrics, register
agreement and Counterfactual Update Selectivity (\CUS) with the move-swap and
conflicting-continuation selectivities, are defined in \Cref{sec:assay}. An ablation masks a
candidate component during the forward pass and rescores \CUS; the matched control routes
the same component from the visible coordinate rather than the hidden coordinate; and a
restoration recovers the component's contribution from the clean edited run. The following criteria determine when a component, or a set of components, mediates the update.

For a \emph{single localized mediator} we require a
reduction of at least 80 percent with the matched control preserved (reduction at most 20
percent), a restoration that recovers at least 70 percent of the lost selectivity, and ordinary
unedited behavior intact. When no single component meets this bar, the update may still be
mediated by a distributed pathway, a small fixed set of components acting together. For a
distributed pathway the appropriate threshold is lower, because the effect is shared:
a joint ablation of a fixed edge set counts as \emph{strong} at a reduction of at
least 70 percent with matched controls preserved and restoration successful, and \emph{partial}
between 40 and 70 percent. Applying the 80 percent single-edge bar to a genuinely
distributed pathway would misreport it as a failure.

These percentages are \emph{predeclared conventions}: each is the fraction of the
baseline selectivity that an ablation removes, and the cutoffs were fixed before scoring
rather than derived from any theory. They serve as a decision rule, in the role a
significance level plays for a statistical test, so the choice cannot be tuned after seeing
the data. The conclusions do not turn on their exact values: the \(Q_8\) edge removes
94 percent of the selectivity, well above the 80 percent bar, and the strongest single
\(D_8\) edge removes 24.2 percent, well below it, so both verdicts survive any
reasonable choice of cutoff. The single boundary case is the \(D_8\) route, whose joint
reduction averages 68.7 percent across ten reported splits (95 percent CI 66.9--70.4), below the 70 percent
strong/partial line, so we report the route conservatively as
\emph{partial} rather than strong.

The distributed pathway for \(D_8\)
was tested with a nested ablation of a fixed four-edge set, frozen before
scoring; it reaches the partial tier (\Cref{sec:d8}).

\paragraph{\(D_8\) route selection protocol.} The four-edge set (the L22, L25, and L23
current-hidden-bit edges and the L19 move edge) was selected on the single-ablation candidate scan
together with two selection splits (seeds 0--1), then frozen before the confirmatory evaluation. The
confirmatory evaluation added eight independently sampled held-out splits (seeds 2--9); we
fixed \(N = 10\) in advance and report all ten splits, summarized by the means in \Cref{fig:cutwire}
and released in full with the code. No edge was
added, removed, reordered, or rescored after observing the ten-split outcomes. Because two of the
ten splits informed selection, the eight held-out splits alone average 68.4 percent,
indistinguishable from the ten-split mean, so the estimate is not inflated by selection.

We localize the update with the same single-ablation scan in both groups: a fixed set of
38 candidate components on the path to the next-state readout (attention edges and low-rank
adapter bands) each ablated in turn under the register edit. The two groups separate across the
whole search space (\Cref{fig:landscape}): in \(Q_8\) one component, the layer-22 current-state
hidden-bit edge, removes 94 percent of the selectivity while every other candidate removes at most
15 percent; in \(D_8\) the same edge is the strongest candidate but removes only 24
percent, and the rest are far lower, with no single mediator.

\begin{figure}[!htbp]
\centering
\includegraphics[width=\linewidth]{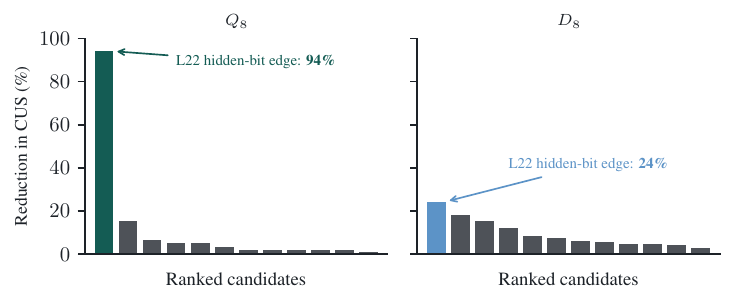}
\caption{\textbf{The update is concentrated in \(Q_8\) and distributed in \(D_8\).} Bars show the
leading candidates from the fixed 38-candidate scan, ranked by the reduction in Counterfactual
Update Selectivity (\CUS) when each is ablated alone; only the leading candidates are plotted, and the remaining lie in the near-zero tail (the full 38-candidate ranking is released with the code). In \(Q_8\) one component dominates (the layer-22
current-state hidden-bit edge, teal; 94 percent), and the next-largest removes only 15 percent.
In \(D_8\) the same edge is the strongest candidate (blue) but removes only 24 percent, and the
remaining candidates decline gradually, with no isolated mediator.}
\label{fig:landscape}
\end{figure}

\FloatBarrier
\subsection{A single attention edge mediates the \texorpdfstring{\(Q_8\)}{Q8} update}

\label{sec:localize-q8}

In \(Q_8\), a single attention edge mediates the counterfactual update. Scanning a fixed candidate set on the path to the
next-state readout, a single attention edge stands out, in the attention-head circuit terms
of the transformer-circuits analyses \cite{elhage2021framework,olsson2022induction}: the edge
from the current-state hidden coordinate to the next-state readout at layer~22. Its ablation removes about
94 percent of the baseline selectivity, driving \CUS\ from 0.595 to 0.035; its
restoration recovers \CUS\ to 0.595; and the matched control edge, routed from the visible
coordinate to the same readout, leaves \CUS\ near 0.59. Ordinary unedited behavior is
preserved. This edge is the isolated winner of the scan: the next-best candidate removes
about a sixth as much (\Cref{fig:landscape}). The localization reproduces on an independent
data split, where the same edge is again the isolated winner and again meets the full
criterion, so it is not an artifact of one split. \Cref{fig:cutwire} shows the ablation,
control, and restoration for the \(Q_8\) edge and for the \(D_8\) route.

\begin{figure}[!htbp]
\centering
\includegraphics[width=0.72\linewidth]{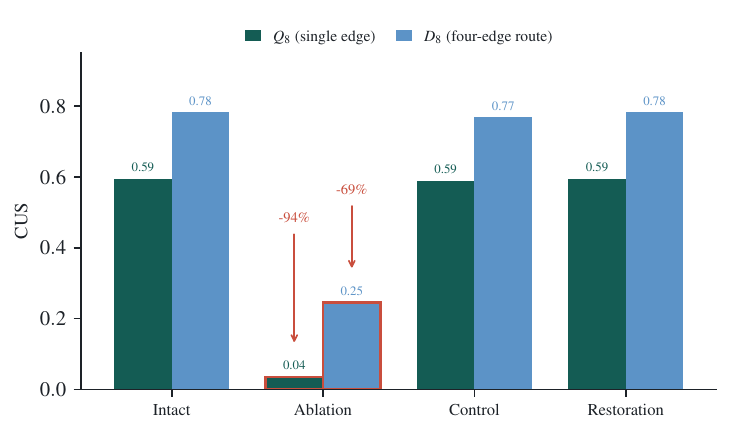}
\caption{\textbf{Ablation and restoration validate the update routes.} Counterfactual
Update Selectivity (\CUS) for \(Q_8\) (teal; the single layer-22 edge) and \(D_8\) (blue; the
frozen four-edge route, mean over ten reported splits, eight held-out confirmatory) under the
intact edit, ablation of the route, a matched control, and restoration. Ablating the route
drives \CUS\ to near zero in both groups (a 94 percent reduction in \(Q_8\) and 68.7
percent in \(D_8\), highlighted), while the matched control and the restoration hold near the
intact baseline.}
\label{fig:cutwire}
\end{figure}

\begin{table}[!htbp]
\centering
\small
\caption{\textbf{Localization in \(Q_8\) replicates across splits.} Ablating the
layer-22 edge that carries the hidden coordinate into the next-state readout removes more
than 90 percent of the selectivity on both splits; the matched control edge is preserved;
the restoration recovers the selectivity. Ordinary unedited behavior, \(\mathbb{P}(\hnext =
h(\ms))\), is preserved at \(0.82\) and \(0.79\). The edge is the isolated winner of the
candidate scan; the next-best candidate removes \(0.090\) of selectivity on split~1 and
\(0.110\) on split~2.}
\label{tab:q8loc}
\begin{tabularx}{0.96\linewidth}{Xcccc}
\toprule
\textbf{\(Q_8\) localization (layer-22 edge)} & baseline & ablation & matched control & restoration \\
\midrule
\CUS, split 1 & \(0.595\) & \(0.035\) & \(0.59\) & \(0.595\) \\
\CUS, split 2 (replication) & \(0.525\) & \(-0.050\) & \(0.53\) & \(0.525\) \\
relative reduction & --- & \(0.94\) / \(1.10\) & \(\le 0.02\) & restored \(\ge 0.99\) \\
\bottomrule
\end{tabularx}
\end{table}

\FloatBarrier
\subsection{The frozen \texorpdfstring{\(D_8\)}{D8} route and its replication across splits}

\label{sec:d8}

In \(D_8\) the register replicates: editing it redirects the computation at 0.91 register
agreement, with a baseline \CUS\ of 0.825, and the move-swap and conflicting-continuation
controls hold. The update route, however, is distributed rather than concentrated in a single
edge. No individual candidate
passes the single-edge criterion, and the layer-22 edge that mediates the \(Q_8\) update
removes only about a quarter of the \(D_8\) selectivity (\Cref{fig:spec-b}). We therefore test a fixed set of four edges, selected on the candidate scan and two selection
splits, then frozen, in which the current-state hidden bit reaches the
next-state readout through layers 22, 25, and 23 and the move token enters at layer 19.
Ablating the set removes a mean of 68.7 percent of the selectivity across ten reported
splits, eight of them independent held-out confirmatory splits (95 percent CI 66.9 to 70.4):
a replicated partial localization, consistent
across splits but below the strong threshold met in \(Q_8\). Across all ten splits both
matched control sets preserve the effect, the clean restoration returns it to baseline within reported precision, and ordinary
behavior stays intact (\Cref{tab:d8scan,tab:d8-ten-splits}).

The four-edge route is also unusually strong among same-size alternatives. Among 1000 random
four-edge sets drawn from the 32 literal attention edges in the same candidate landscape, the
median \CUS\ removed is 0.055 and the 95th percentile is 0.254; the frozen route, which
removes 0.532 on the held-out splits, lies at the 99.6th percentile (\Cref{app:d8null}). A
subset analysis shows the route is distributed rather than single-edge: the strongest individual
edge (layer 22) removes only 0.212, the full route removes 0.532, and the full route
exceeds the sum of its singleton effects. This is a joint route with a dominant late current-state edge,
not four equal mediators (\Cref{app:d8subset}).

\begin{table}[!htbp]
\centering
\small
\caption{\textbf{The \(D_8\) update is partially localized to a distributed pathway.} A nested
ablation of the fixed four-edge set \(E_4\) (selected on the candidate scan and two selection splits, then frozen) removes
a mean of \(68.7\) percent of the selectivity across ten reported splits (eight held-out confirmatory; 95 percent CI
\(66.9\)--\(70.4\); the accumulation shown is the seed-0 selection split): a replicated partial localization that does not meet the strong
single-edge criterion.
The current-state hidden bit routes through layers 22, 25, and 23 into the next-state readout, and
the move token enters at layer 19. Across all ten splits both matched controls preserve the
selectivity (the quotient-source and off-target destination controls stay near \(0.82\)), the
clean restoration recovers it, and ordinary unedited behavior is intact
(\(\mathbb{P}(\hnext = h(\ms)) = 0.88\)). No single edge passes the single-edge criterion (the
largest, L22, removes only \(24\) percent), so the \(D_8\) update is distributed where the
\(Q_8\) update is concentrated.}
\label{tab:d8scan}
\begin{tabularx}{0.96\linewidth}{Xcc}
\toprule
\textbf{\(D_8\) cumulative ablation} (baseline \CUS\ \(= 0.825\)) & ablation \CUS\ [95\% CI] & reduction \\
\midrule
\(E_1\): L22 current hidden bit & \(0.625\ [0.550,\,0.700]\) & \(24.2\%\) \\
\(E_2\): \(+\) L19 move & \(0.575\ [0.495,\,0.655]\) & \(30.3\%\) \\
\(E_3\): \(+\) L25 current hidden bit & \(0.555\ [0.470,\,0.635]\) & \(32.7\%\) \\
\(\boldsymbol{E_4}\): \(+\) L23 current hidden bit & \(\mathbf{0.265\ [0.170,\,0.360]}\) & \(\mathbf{67.9\%}\) \\
\bottomrule
\end{tabularx}
\end{table}

\begin{table}[!htbp]
\centering
\scriptsize
\setlength{\tabcolsep}{3pt}
\caption{\textbf{Ten-split \(D_8\) frozen-route results.} Seeds \(0\)--\(1\) are the two
selection splits; seeds \(2\)--\(9\) are held-out confirmatory. The frozen four-edge route
has mean reduction \(68.69\%\) with split-bootstrap 95\% CI \([66.88\%,\,70.35\%]\). The clean
restoration equals the reported baseline \CUS\ in all ten splits; the quotient-source and
off-target controls preserve selectivity in every row.}
\label{tab:d8-ten-splits}
\begin{tabular}{clrrrrrrr}
\toprule
Seed & Split class & Base \CUS & Ablated \CUS & Quotient ctrl & Off-target ctrl & Restoration & Behavior & Reduction \\
\midrule
\(0\) & selection & \(0.825\) & \(0.265\) & \(0.835\) & \(0.820\) & \(0.825\) & \(0.8750\) & \(67.88\%\) \\
\(1\) & selection & \(0.790\) & \(0.225\) & \(0.790\) & \(0.755\) & \(0.790\) & \(0.8875\) & \(71.52\%\) \\
\(2\) & held-out confirmatory & \(0.780\) & \(0.230\) & \(0.790\) & \(0.775\) & \(0.780\) & \(0.9025\) & \(70.51\%\) \\
\(3\) & held-out confirmatory & \(0.725\) & \(0.230\) & \(0.725\) & \(0.695\) & \(0.725\) & \(0.8800\) & \(68.28\%\) \\
\(4\) & held-out confirmatory & \(0.775\) & \(0.245\) & \(0.785\) & \(0.765\) & \(0.775\) & \(0.8825\) & \(68.39\%\) \\
\(5\) & held-out confirmatory & \(0.795\) & \(0.235\) & \(0.795\) & \(0.780\) & \(0.795\) & \(0.8800\) & \(70.44\%\) \\
\(6\) & held-out confirmatory & \(0.795\) & \(0.270\) & \(0.805\) & \(0.775\) & \(0.795\) & \(0.8750\) & \(66.04\%\) \\
\(7\) & held-out confirmatory & \(0.785\) & \(0.210\) & \(0.790\) & \(0.770\) & \(0.785\) & \(0.9050\) & \(73.25\%\) \\
\(8\) & held-out confirmatory & \(0.780\) & \(0.290\) & \(0.790\) & \(0.790\) & \(0.780\) & \(0.8800\) & \(62.82\%\) \\
\(9\) & held-out confirmatory & \(0.775\) & \(0.250\) & \(0.785\) & \(0.760\) & \(0.775\) & \(0.8725\) & \(67.74\%\) \\
\bottomrule
\end{tabular}
\end{table}

\FloatBarrier
\subsection{Matched random routes and the contributions of individual edges}
\label{app:d8null}
\label{app:d8subset}

\paragraph{Random four-edge comparisons.}

On the eight held-out confirmatory splits the frozen four-edge route \(E_4\) removes a mean
0.532 \CUS. We compare it to 1000 random four-edge sets drawn from the 32 literal
attention edges in the same candidate landscape; the broad LoRA-band candidates are excluded
because they are not edge-comparable. The null has median 0.055 and 95th percentile 0.254,
and \(E_4\) lies at the 99.6th percentile (\Cref{fig:d8null}). The route is thus unusually
strong among same-size attention-edge sets: it is not the case that any four plausible edges
remove two-thirds of the effect. We do not claim \(E_4\) is the single most destructive four-edge
set (the null maximum, 1.022, exceeds it) only that it is an unusually strong route that also
passes the matched-control and restoration checks of \Cref{sec:d8}. Conditioning the null further on
\(E_4\)'s edge signature (three current-hidden-bit edges and one move edge into the next-state
readout) leaves \(E_4\) at the 99.0th percentile among same-signature random four-edge sets
(median 0.093, 95th percentile 0.336), so the route is unusually strong even against type-
and destination-matched alternatives.

\begin{figure}[!htbp]
\centering
\includegraphics[width=0.82\linewidth]{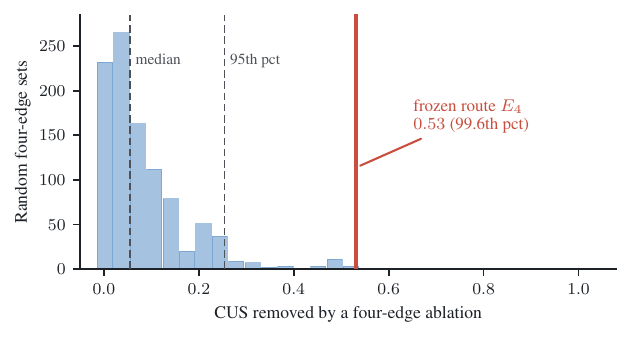}
\caption{\textbf{The \(D_8\) four-edge route is unusually strong among same-size edge sets.}
Distribution of \CUS\ removed by 1000 random four-edge sets drawn from the 32 literal
attention edges in the candidate landscape (broad LoRA-band candidates excluded). The frozen route
\(E_4\) (coral) removes 0.532 and lies at the 99.6th percentile; the null median is
0.055 and the 95th percentile is 0.254. A single random set removes more than \(E_4\), so
the claim is percentile and route validation, not uniqueness.}
\label{fig:d8null}
\end{figure}

\paragraph{Subset and leave-one-out analysis.}

\Cref{tab:d8subset} reports, for each edge in \(E_4\), its effect ablated alone and its
leave-one-out contribution (the drop in the full-route effect when that edge is removed). The
strongest individual edge, the layer-22 current-hidden-bit edge, removes only 0.212 \CUS,
while the full route removes 0.532; no singleton approaches the route. The full route also
exceeds the sum of the singleton effects (0.472), and the layer-22 and layer-23 edges carry
the largest conditional contributions. The \(D_8\) update is therefore distributed and partly
synergistic: a joint route with a dominant late current-state edge, not single-edge mediation and
not four equal independent effects.

\begin{table}[!htbp]
\centering
\caption{\textbf{Subset and leave-one-out anatomy of the \(D_8\) four-edge route}, on the eight
held-out confirmatory splits. ``Alone'' ablates only that edge; the leave-one-out drop is the
reduction in the full-route effect when that edge is removed from \(E_4\). Entries are mean \CUS\
removed.}
\label{tab:d8subset}
\begin{tabularx}{0.86\linewidth}{Xcc}
\toprule
Edge & Alone & Leave-one-out drop \\
\midrule
Layer 22, current hidden bit & \(0.212\) & \(0.346\) \\
Layer 19, move token & \(0.113\) & \(0.021\) \\
Layer 25, current hidden bit & \(0.078\) & \(0.039\) \\
Layer 23, current hidden bit & \(0.069\) & \(0.258\) \\
\midrule
Full route \(E_4\) (all four) & \(0.532\) & --- \\
\bottomrule
\end{tabularx}
\end{table}

\FloatBarrier
\subsection{Register dimensionality and update-route concentration}
\label{app:rank-robustness}

These results separate register compactness from circuit concentration. Trace supervision creates an analogous compact register in both groups, and the route that
updates it is concentrated in one edge in \(Q_8\) and distributed across a small set of edges
in \(D_8\); the same token classes enter the nominated route in both groups. The main mechanistic contrast is the concentration of the update route.

The register itself is low-dimensional. Sweeping the edit rank shows the
counterfactual update emerge and saturate by rank~1 in \(Q_8\) and rank~2 in \(D_8\)
(\Cref{fig:spec-a}), far below the 3584-dimensional residual stream. The rank profile is
consistent with a one-bit register: in \(Q_8\) a single causal direction suffices to hold and
flip it. In \(D_8\) the leading variance direction is not the
causal one, so the rank-1 edit on its own fails and even anti-selects the move-applied target,
while its orthogonal complement still carries the effect; the second direction recovers it,
placing the causal axis in a two-dimensional subspace.

\begin{figure}[!htbp]
\centering
\begin{subfigure}[b]{0.49\linewidth}
\centering
\includegraphics[width=\linewidth]{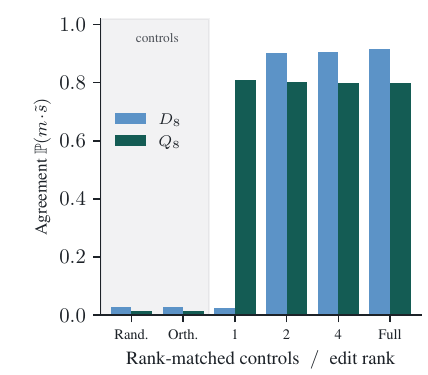}
\caption{Compactness}\label{fig:spec-a}
\end{subfigure}\hfill
\begin{subfigure}[b]{0.49\linewidth}
\centering
\includegraphics[width=\linewidth]{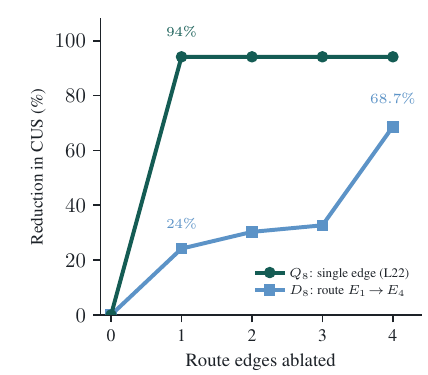}
\caption{Concentration}\label{fig:spec-b}
\end{subfigure}
\caption{\textbf{A compact variable with a concentrated or distributed update route.}
(a) Agreement is near zero for the random and orthogonal controls and reaches its plateau by
rank~1 in \(Q_8\) and rank~2 in \(D_8\), far below the residual dimension, so the causal register is one
to two dimensional; the \(D_8\) rank-1 edit alone fails because its leading direction is
misaligned with the causal one. (b) Cumulative reduction in selectivity as route
edges are ablated one at a time. In \(Q_8\) a single layer-22 edge already reaches the full
94 percent reduction, so the curve steps up at the first edge and saturates (concentrated).
In \(D_8\) no single edge suffices: the reduction accumulates along the frozen four-edge route
(24, 30, 33 percent, on the selection split); the largest increment comes when the fourth
edge is added, reaching a replicated partial 68.7 percent, the ten-split mean (distributed).}
\label{fig:specificity}
\end{figure}

\paragraph{Layer and rank robustness.} The register effect is stable across the trained layer and the
neighboring layer, and a rank sweep shows the edit is already at full strength by rank~1 in
\(Q_8\) and rank~2 in \(D_8\) (random and orthogonal controls stay near zero), so the register
is one to two dimensional. The \(D_8\) rank-1 edit is the lone exception: its leading direction
is misaligned with the causal one, so it fails on its own while its orthogonal complement still
carries the effect. The localization holds at the reported layer on both data splits
(\Cref{tab:q8loc}).

\FloatBarrier

\section{Function Vector construction and target-activation comparisons}
\label{app:steering-methods}
\suppressfloats[t]

The state-tracking comparison asks whether Function Vector steering directly writes the identified register. The lexical experiments ask whether effective steering recreates the matched target activation at the intervention site. The constructions for these two settings are reported separately below, followed by the lexical scoring protocol and complete comparison results.

\FloatBarrier
\subsection{Function Vector construction for \texorpdfstring{\(Q_8\)}{Q8}}
\label{app:fv-steering-find}

We applied Function Vector steering to the \(Q_8\)-adapted
\texttt{Qwen2.5-Coder-7B}. Vectors extracted from demonstration prompts
transferred poorly to native \(Q_8\) trace continuations, so we estimated head
activations directly from those traces. At the current phase token, we estimated each attention
head's mean output after the \(W^O\) projection from \(1{,}024\) \(Q_8\) trace
examples, with \(128\) examples per state.

On \(256\) separate, state-balanced
examples, we ranked heads by the average increase in the probability of the
correct next-phase token when the head's target-minus-source mean was added at
layer~12. The readout was conditioned on the correct next-region prefix. We
selected the five heads with the largest positive average effects and used the
same head set for every state.

For each region, we summed the selected heads' mean outputs separately for phases
\texttt{P} and \texttt{R}, then subtracted the \texttt{P} sum from the \texttt{R}
sum to obtain the regional steering vector. To change \texttt{P} to \texttt{R},
we added eight times this vector to layer-12 \texttt{resid\_pre} at the current
phase token; for the reverse edit, we subtracted the same scaled vector. The head
count and intervention scale were selected on \(256\) development examples and
frozen before evaluation on \(400\) held-out examples. Mean estimation, head
ranking, development, and held-out evaluation used disjoint trace records.

\FloatBarrier
\subsection{Lexical tasks, vector construction, and comparison interventions}

The steering experiments use \texttt{Qwen/Qwen2.5-7B} and
\texttt{meta-llama/Llama-3.1-8B} in BF16 without quantization or chat templates. Task pairs and
the Function Vector construction follow Todd et al. \cite{todd2024fv}. Ten demonstrations are
rendered as \texttt{Q: input} and \texttt{A: output}; the query ends immediately after the
\texttt{A:} prefix. Demonstration outputs are either correct or deterministically shuffled while
preserving token alignment at the final answer position.

Inputs and outputs are split without replacement into 45 prompts for constructing the steering
direction, 32 prompts for calibration, and 128 prompts for the initial evaluation. The comparison
with the model's target activations uses 64 additional prompts per selected model-task setting.
These prompts do not overlap with the construction, calibration, or initial evaluation data. Each
target and shuffled prompt uses the same ten demonstrations and query.

Attention heads are ranked by average indirect effect. For each selected head, we replace its input
to the output projection at the final prompt token with the corresponding average from target-task
prompts while the model receives a shuffled prompt. The outcome is the probability of the first
target token. The Qwen direction combines heads \((17,15)\), \((18,23)\), and \((16,13)\); the
Llama direction uses head \((13,27)\). Their projected outputs are averaged over the 45 construction
prompts. Separate calibration data select one intervention layer and coefficient for each model,
yielding block 7 and coefficient 15 for Qwen, and block 8 and coefficient 10 for Llama.

The final comparison uses one effective setting per model:
\begin{center}
\small
\begin{tabular}{@{}llll@{}}
\toprule
Model & Task & Intervention block & Coefficient \\
\midrule
Qwen2.5-7B & present-to-past & 7 & 15 \\
Llama-3.1-8B & English-to-Spanish & 8 & 10 \\
\bottomrule
\end{tabular}
\end{center}

Each evaluation prompt has a deterministic alternative answer from the same output set, with the
target-first and alternative-first order balanced 32/32. We score complete target-plus-EOS and
alternative-plus-EOS log probabilities under the same prefix. In the all-layer target condition,
we replace the activation at the answer position after every remaining block with the matched
activation from the target run. In the one-layer target condition, we make this replacement only
after the intervention block. Additive conditions use the Function Vector, the same direction at
the norm of the prompt's natural activation change, the prompt's natural target direction at the
Function Vector norm, a held-out average of 45 target-minus-shuffled activation changes, the other
task's Function Vector, or a deterministic random direction.

\FloatBarrier
\subsection{Scoring and normalization}

\label{app:steering-scoring}

For each evaluation prompt \(i\), we fix a target answer and an alternative answer and score the
log probabilities of their complete strings, including the end-of-sequence token. Under condition
\(c\), the target-minus-alternative margin is
\[
M_i(c)
=
\log p_c\!\left(y_i^{\mathrm{target}}\mid x_i\right)
-
\log p_c\!\left(y_i^{\mathrm{alternative}}\mid x_i\right),
\]
where each answer log probability is the sum of its autoregressive token log probabilities. The
paired gain from intervention \(c\), relative to the same unedited shuffled prompt, is
\[
\Delta_i(c)=M_i(c)-M_i(\mathrm{unedited}).
\]

To compare effects across models, we normalize by the repeated correct-prompt patch, abbreviated as
the \emph{all-layer patch}. This reference replaces the final-token hidden vector with its matched
correct-prompt value after every remaining block. Its prompt-level gain is
\[
G_i=M_i(\mathrm{all\mbox{-}layer\ patch})-M_i(\mathrm{unedited}),
\]
and the normalized effect is
\[
E(c)
=
\frac{\frac{1}{n}\sum_{i=1}^{n}\Delta_i(c)}
     {\frac{1}{n}\sum_{i=1}^{n}G_i}.
\]
Thus, \(E(c)=0\) indicates no average improvement over the unedited shuffled prompt, whereas
\(E(c)=1\) indicates the same average margin gain as the all-layer patch. Negative values indicate
movement toward the alternative answer. Values above \(1\) are possible because the all-layer
patch is a reference rather than an upper bound. This score compares behavioral effects and does
not measure similarity between the hidden vectors produced by different interventions.

We report paired 95\% bootstrap intervals from \(2{,}000\) prompt-level resamples. Within each
resample, we recompute both the intervention's mean paired gain and the all-layer reference gain,
preserving the within-prompt comparison with the unedited condition. The all-layer patch has
positive average gain in every bootstrap resample, so the normalization denominator is always
well-defined. The task, model, intervention layer and coefficient, prompt sets, and scoring
analysis were fixed before the outcomes in this comparison were examined.

\FloatBarrier
\subsection{Target-activation patches and equal-norm comparison results}

\label{app:steering-full-table}

\begin{table}[!htbp]
\caption{\addedtext{\textbf{Target-activation patches and comparison directions.} Entries are normalized
effects with paired 95\% bootstrap intervals. Patching the target activation through all later
layers selects the target answer on every prompt in both models.}}
\centering
\small
\resizebox{\linewidth}{!}{%
\begin{tabular}{@{}lcc@{}}
\toprule
Condition & Qwen2.5-7B, present-to-past & Llama-3.1-8B, English-to-Spanish \\
\midrule
Target activations at all later layers & 1.000 & 1.000 \\
Mean steering vector & 0.544 [0.384, 0.720] & 0.824 [0.596, 1.084] \\
Target activation at one layer & 0.014 [0.005, 0.023] & -0.015 [-0.041, 0.005] \\
Natural target direction at steering norm & 0.612 [0.391, 0.856] & -0.101 [-0.325, 0.094] \\
Mean steering at natural-change norm & 0.0073 [-0.0003, 0.0155] & 0.0166 [-0.0021, 0.0349] \\
Held-out average activation change & 0.613 [0.332, 0.933] & 0.085 [-0.085, 0.244] \\
Different-task steering vector & 0.665 [0.505, 0.841] & -0.028 [-0.269, 0.192] \\
Random direction & 0.175 [0.090, 0.270] & -0.316 [-0.501, -0.174] \\
\bottomrule
\end{tabular}}
\end{table}

The mean gains from patching target activations through all later layers are 4.412 nats
[3.980, 4.853] for Qwen and 2.639 nats [2.206, 3.099] for Llama. The Function Vector norms are
210.80 and 20.22; the median natural activation-change norms are 2.74 and 0.78.

\paragraph{Validation of the all-layer reference.}
The all-layer patch serves as the positive reference used to normalize the effects of the other
interventions. In each model-task setting, replacing the final-token hidden vector with its
correct-prompt value after every remaining block causes the model to select the target answer on
all 64 evaluation prompts. The patch also has positive average gain in every bootstrap resample,
so the normalization denominator is always well-defined.

This result establishes that the correct-prompt hidden vectors contain sufficient information to
recover the target behavior when they are repeatedly restored throughout the remaining
computation. It does not show that the correct-prompt activation at any single block is sufficient,
nor does it imply that the all-layer patch is an upper bound on the effect of other interventions.
We therefore use it only as a successful causal reference against which to express the average
behavioral gains of the local interventions.

\paragraph{Direction specificity.}
Because the calibrated steering vectors are much larger than the models' natural activation
changes, we compare them with other directions rescaled to the same norm
(\Cref{fig:steering-comparison}b). In Llama, the target task's steering vector has normalized
effect 0.824. The prompt's natural target direction has effect \textminus{}0.101, a steering vector
for another task has effect \textminus{}0.028, a held-out average activation change has effect 0.085,
and a random direction has effect \textminus{}0.316. Among the directions tested, the Llama effect is
therefore specific to the constructed steering direction.

Qwen behaves differently. Its natural target direction, held-out average activation change, and
steering vector for another task have effects 0.612, 0.613, and 0.665, compared with
0.544 for the target task's steering vector. A random direction is weaker at 0.175, so
Qwen is not equally sensitive to every perturbation, but the target steering vector is not unique
among structured directions of the same norm. Complete paired bootstrap intervals are reported in
\Cref{app:steering-full-table}.

Directional specificity strengthens the interpretation of Llama's behavioral response, but it
does not establish state-like use. The one-block target-activation patch and natural-scale steering
tests remain near zero, so even the direction-specific Llama effect does not show that the vector
recreates the model's natural target activation at the tested layer.

\FloatBarrier

\section{Steering persistence, output binding, and supplementary controls}
\label{app:steering-diagnostics}
\suppressfloats[t]

These experiments test explanations of the steering effect beyond its immediate scale and direction: whether it becomes independent of the added vector, whether it follows reassigned output labels, and whether a measured local sensitivity accounts for it.

\FloatBarrier
\subsection{Removing the added vector at later layers}

This experiment asks whether an additive steering edit is quickly converted into a
self-sustaining internal change. We add the vector at its usual layer and then subtract the same
vector at a later layer, leaving all other changes produced by the intervening computation intact.
If the model has already converted the edit into another task-specific representation, the
behavioral effect should remain after the original vector is removed.

For each model, we construct eight steering directions across four lexical tasks and data folds. Seven
directions per model produce a clear steering effect relative to matched comparisons and are
carried forward. In all fourteen cases, removing the vector eliminates the task-specific effect
until late in the network. The first layers where much of the effect remains are blocks 22--24 in
Qwen and 18--25 in Llama. Matched comparison directions also begin to affect the answer at blocks
9--15 and 8--10, respectively. Because task specificity and comparison-direction effects emerge at
similar depths, the experiment does not isolate a task-specific representation that becomes
self-sustaining early in the network.

\FloatBarrier
\subsection{Counterfactual output binding}

We assign the correct answer to either of two arbitrary output labels while holding the underlying
task fixed. For the paired assignments, let \(g_0\) and \(g_1\) be the edit-induced changes in the
physical label-0-minus-label-1 margin. We decompose
\[
S=(g_0-g_1)/2,\qquad T=(g_0+g_1)/2,
\]
where \(S\) follows the semantic role assigned by the prompt and \(T\) remains tied to the physical
labels.

\begin{table}[!htbp]
\centering
\small

\caption{\textbf{The steering edit does not follow a counterfactual answer mapping.}
The component that follows the assigned semantic role is negative in both models, while an effect
tied to the physical output labels remains.}
\begin{tabular}{@{}lrrrr@{}}
\toprule
Model & Directions & \(S\) & 95\% CI & \(T\) \\
\midrule
Qwen2.5-7B & 7 & -0.315 & [-0.391, -0.236] & -1.094 \\
Llama-3.1-8B & 5 & -0.250 & [-0.278, -0.218] & 0.225 \\
\bottomrule
\end{tabular}
\end{table}

Without steering, Qwen follows the assigned answer mapping on 93.3\% of target pairs and Llama on
70.3\%. Restricting the analysis to pairs solved under both mappings makes \(S\) more negative in
both models, so Llama's lower baseline agreement does not explain the sign of the main result. The
vector therefore does not behave like an abstract ``correct answer'' signal that can be rebound to
whichever output label the prompt designates as correct.

\FloatBarrier
\subsection{Supplementary local-sensitivity comparison}

This supplementary local-sensitivity test is inconclusive because BF16 conversion misses the prespecified tolerance by approximately \(10^{-4}\); we report it for completeness but do not use it to support the main claim.

We compare the Function Vector with a nearly orthogonal control direction constructed to have the
same measured first-order effect on the target-minus-alternative margin. We also project the
Function Vector to remove that measured first-order component and apply the unchanged vector only
after the final decoder block.

\begin{table}[!htbp]
\centering
\small

\caption{\textbf{Supplementary local-sensitivity comparison.}
Effects are normalized to the paired gain from patching target activations through all later
layers. The matching check described below misses its specified tolerance, so these values are
included for completeness and are not used to support the main conclusion.}
\begin{tabular}{@{}lcc@{}}
\toprule
Condition & Qwen2.5-7B & Llama-3.1-8B \\
\midrule
Mean steering vector & 0.544 [0.390, 0.714] & 0.824 [0.593, 1.072] \\
Sensitivity-matched control & 0.097 [-0.090, 0.286] & -0.137 [-0.346, 0.062] \\
Sensitivity-removed steering vector & 0.545 [0.391, 0.726] & 0.796 [0.578, 1.056] \\
Steering only after the final block & -0.032 [-0.089, 0.026] & 0.057 [-0.008, 0.126] \\
\bottomrule
\end{tabular}
\end{table}

The near-orthogonality check is satisfied on all 128 prompts. After BF16 conversion, however, the
relative norm-matching errors reach \(9.81\times10^{-5}\) in Qwen and \(1.24\times10^{-4}\) in
Llama, above the specified \(10^{-5}\) tolerance. We do not relax this criterion or rerun the
experiment. The observed pattern is consistent with the measured scalar derivative and an
unchanged final-layer readout being insufficient explanations, but the failed numerical check
prevents a stronger conclusion and the experiment does not identify the downstream computation.

\FloatBarrier

\section{Interpretation, scope, and related work}
\label{app:discussion-details}
\suppressfloats[t]

The distinction between behavioral control and computational state determines how we interpret both sets of experiments. We collect that interpretation, its limits, and the connections to prior work here.

\FloatBarrier
\subsection{What the intervention evidence establishes}

\begin{table}[!htbp]
\caption{\textbf{What different intervention tests establish.} Changing an answer demonstrates
control. Interpreting the edit as computational state additionally requires showing that later
operations use the edited value as the proposed variable.}
\label{tab:causal-roles}
\centering
\small
\begin{tabularx}{\linewidth}{@{}lXX@{}}
\toprule
Interpretation & Test & What the result shows \\
\midrule
Direct output effect & Apply the edit at the final output boundary. & The intervention directly
changes the reported answer. \\
Downstream control & Apply the edit before later layers and test whether behavior changes. & Later
computation responds to the intervention. \\
Computational state & Write a counterfactual value, vary the next operation, and test the resulting
future state. & The model uses the edited value according to the proposed variable's update rule. \\
\bottomrule
\end{tabularx}
\end{table}

\paragraph{Registers and update routes.}

The identified register is low-dimensional, although the route that updates it need not be
equally concentrated. The causal effect saturates at rank one in \(Q_8\) and rank two in \(D_8\).
In \(Q_8\), ablating a single attention edge removes about 94\% of the effect, whereas
ablating the identified four-edge route removes 68.7\% in \(D_8\). The dimensionality of a
stored variable and the concentration of the computation that reads and updates it are therefore
separate properties.

The counterfactual branch also persists during free-running generation, cleanly in \(D_8\) and
more partially in \(Q_8\), and the core register result replicates in a second model family. The
circuit analysis itself is established only in the primary model. Full free-running, replication,
and localization results appear in \Cref{app:freerun,app:second-family,app:localize}.

\paragraph{Scale and direction of steering.}

Replacing the final-token residual once with the exact residual from the correct prompt produces
less than 2\% of the all-layer reference effect in either model. Rescaling the Function Vector
on each prompt to the size of the natural correct-versus-shuffled change is similarly ineffective.
At its calibrated scale, the vector is effective, but its norm is 77 times the median natural
change in Qwen and 26 times that change in Llama.

Successful steering at these sites is therefore not well described as restoring the residual that
the model reaches on the matched correct prompt. Instead, the intervention moves the
shuffled-prompt residual by a large, externally imposed displacement. This is the narrow, measured
sense in which the steered residual is unnatural: the comparison concerns the immediate
displacement at the intervention site, not the likelihood of that point under the full activation
distribution or the geometry of its subsequent trajectory.

The equal-norm controls show that scale alone does not determine the effect. In Llama, only the
target Function Vector produces a clearly positive target-directed effect among the equally large
directions tested. Qwen is less selective, responding to several structured directions but more
weakly to a random direction. Direction can therefore matter, but this does not establish that the
intervention recreates a natural target activation or behaves like a register write. Full numerical
results appear in \Cref{app:steering-full-table,app:steering-diagnostics}.

\paragraph{Persistence and output binding.}

The cancellation experiment asks whether the initial steering displacement is quickly converted
into a separate, self-sustaining task state. We add the vector at its usual layer and subtract it
at progressively later layers, leaving the intervening computation intact. Early subtraction
eliminates the task-specific effect, even though several layers have already processed the edited
residual. Much of the effect survives only when subtraction is delayed until late in the network.
The experiment therefore shows that the imposed displacement remains causally important under
this test, but it does not identify how later layers use that displacement.

The answer-remapping experiment rules out another simple account. When the underlying task is
held fixed but the output labels are reassigned, part of the steering effect remains tied to the
same literal output label rather than following whichever label the prompt designates as correct.
The vector is therefore not simply supplying an abstract ``correct answer'' signal. Detailed
cancellation and answer-remapping results appear in \Cref{app:steering-diagnostics}.

\FloatBarrier
\subsection{Scope and limitations}

The state-tracking experiments use explicit traces in controlled \(Q_8\) and \(D_8\) tasks rather
than unrestricted natural-language reasoning. The circuit analysis is established only in the
primary model, although the central intervention result replicates in a second model family.

The steering experiments examine one selected task and layer per model, both chosen because
conventional steering was effective. Model, task, layer, and steering scale therefore vary
together, so the specificity difference should not be treated as a general contrast between Qwen
and Llama.

The one-block patch replaces only the final prompt-token residual, while earlier token activations
remain those of the shuffled prompt. Its failure rules out this local restoration account, not a
task state represented at another position or layer, distributed across several activations, or
constructed later in the network. Our use of \emph{unnatural} is similarly local: we compare the
immediate displacement at the intervention site with the matched natural change, but do not
estimate the full distribution of natural residuals or follow the steered residual geometrically
through every remaining layer.

Finally, the cancellation experiment shows that subtracting the imposed vector continues to
remove the measured effect until late in the network, but not how later layers use the
perturbation. The supplementary local-sensitivity comparison also misses its prespecified
numerical tolerance by approximately
\(10^{-4}\). We report it for completeness but do not use it to support the main conclusion.

\FloatBarrier
\subsection{Extended related work}

\label{app:related-work}

\paragraph{Activation steering and task vectors.}
Activation Addition, representation engineering, inference-time intervention, and Function Vectors
show that adding directions to internal activations can control language-model behavior
\cite{turner2023actadd,zou2023repe,li2023iti,todd2024fv}. Function Vectors identify influential
attention heads and average their outputs to elicit a task. Recent work further shows that
activations taken just before the model performs a behavior are particularly effective sources for
steering vectors \cite{ye2026sources}. We ask a different question: when such a vector works, has
it recreated an activation that the model naturally computes and continues from? Comparing the
steering vector with the model's own target activation separates behavioral control from this
stronger interpretation.

\paragraph{Identifiability and reusable interfaces.}
Behaviorally equivalent steering vectors need not identify a unique internal representation
\cite{venkatesh2026nonidentifiability}. The Hidden APIs framework uses multiple possible future operations to
discover internal interfaces that can be transplanted and reused without assuming a known latent
label \cite{ma2026hiddenapis}. Our register experiment uses a complementary source of evidence: the
state variable and transition rule are known by construction, so each counterfactual edit has a
different predicted consequence under each next move. The activation-steering experiments then
show why changing the answer alone does not provide the same evidence.

\paragraph{Causal abstraction and activation patching.}
Distributed alignment search and interchange interventions test whether model variables implement
a proposed high-level computation \cite{geiger2024das,geiger2021abstractions}. Activation patching
and causal tracing localize information by exchanging activations between runs
\cite{meng2022rome}. We compare several interventions that can all change behavior but support
different conclusions: replacing one activation, replacing the sequence of activations through
later layers, adding a fixed direction, and writing a counterfactual register value. The comparison
asks what later computation can do with the edited activation rather than treating every
successful patch as the same kind of explanation.

\paragraph{Scratchpads, state tracking, and world models.}
Scratchpads and chain-of-thought can improve multi-step behavior
\cite{nye2021scratchpad,wei2022cot}, but emitted text need not function as a variable in the
model's later computation. Finite-state analyses identify decodable states in prompted
Transformers \cite{zhang2025fsa}, while entity and belief-tracking work localizes mechanisms that
bind and retrieve state from context \cite{prakash2024entity,prakash2025lookbacks}. Emergent
world-state representations have also been found in self-supervised sequence models
\cite{li2023othello,nanda2023othello}. We intervene on the supervised trace token and ask whether
the model runs the known update rule on the value written there.

\paragraph{Geometry and causal use.}
Linear and nonlinear feature geometry, cyclic representations, and manifold steering provide
useful descriptions of model representations and behavior
\cite{engels2024linear,nanda2023grokking,kantamneni2025trig,feucht2026calculator,wurgaft2026manifold}.
Our result does not conflict with useful steering geometry. It asks for additional causal evidence
before interpreting a direction as computational state: does the model's later computation use the
edited value according to the proposed variable's semantics?

\FloatBarrier

\endgroup



\end{document}